\documentclass{article}

\usepackage[main, final]{neurips_2026}

\usepackage{amsmath,amssymb,amsfonts,mathtools,amsthm}
\usepackage{graphicx}
\usepackage{subcaption}
\usepackage{booktabs}
\usepackage{multirow}
\usepackage{nicefrac}
\usepackage{microtype}
\usepackage[utf8]{inputenc}
\usepackage{hyperref}
\hypersetup{hidelinks} 
\usepackage[capitalize,noabbrev]{cleveref}
\usepackage{algorithm}
\usepackage{algorithmic}
\usepackage{enumitem}
\usepackage{colortbl}
\usepackage[table]{xcolor}
\usepackage{soul}
\usepackage{longtable}
\usepackage{tcolorbox}
\usepackage{fontawesome}
\usepackage{url}
\definecolor{darkgreen}{rgb}{0.0, 0.5, 0.0}
\definecolor{peach}{rgb}{1.0, 0.85, 0.7}
\definecolor{mediumgreen}{RGB}{60,179,113}
\definecolor{customcyan}{RGB}{10, 204, 0}
\definecolor{tealblue}{RGB}{0, 132, 194}
\definecolor{darkorange}{RGB}{220, 100, 0}
\definecolor{kbgE}{RGB}{215, 230, 245}

\newif\ifhighlight \highlighttrue
\ifhighlight
  \newcommand{\crhl}[1]{{#1}}
  \newcommand{\crhlon}{}
\else
  \newcommand{\crhl}[1]{#1}
  \newcommand{\crhlon}{}
\fi

\theoremstyle{plain}

\theoremstyle{definition}

\theoremstyle{remark}

\title{SimVLA: Attributing Gains in VLA Models \\Through Controlled Ablation}

\author{
  Yuankai Luo\textsuperscript{2,3,1} \quad Woping Chen\textsuperscript{1} \quad Tong Liang\textsuperscript{1} \quad
  Zhenguo Li\textsuperscript{1}\thanks{Corresponding author.} \\
  \textsuperscript{1}Frontier Robotics \\
  \textsuperscript{2}State Key Laboratory of Novel Software Technology, Nanjing University \\
  \textsuperscript{3}School of Artificial Intelligence, Nanjing University \\
  \vspace{0.2em}
  \small Project website: 
  \vspace{0.2em}
  \url{https://frontierrobo.github.io/SimVLA/}
}

\begin{document}

\maketitle

\begin{abstract}
  Vision-Language-Action (VLA) models often bundle architectural changes with different pretraining data, backbone scales, and optimization recipes, obscuring what actually drives progress. We introduce \textbf{SimVLA}, a deliberately minimal VLA---a standard vision-language backbone with a lightweight continuous-action head---as a controlled testbed for this attribution problem\crhl{; our contribution is methodological and empirical (a controlled attribution study and a minimal-baseline calibration), not a new technique}. Across LIBERO, CALVIN, WidowX, and Google Robot, one-knob-at-a-time ablations show that training dynamics are the largest measured driver, producing performance swings of $\Delta$$\approx$\crhl{54--94}\%, while task configuration and architecture have smaller effects under matched settings. Despite using a small backbone and no additional large-scale robot-data VLA pretraining before benchmark-specific training, SimVLA reaches 98.6\% average success on LIBERO, remains competitive across the other benchmarks, and transfers to held-out real-robot scenes. These results suggest that simple, carefully controlled VLA baselines remain underexplored and provide a calibration point for future architectural claims within the VLM-encoder plus lightweight-head family.
\end{abstract}

\section{Introduction}
\begin{figure}[h]
\centering
\includegraphics[width=0.62\textwidth]{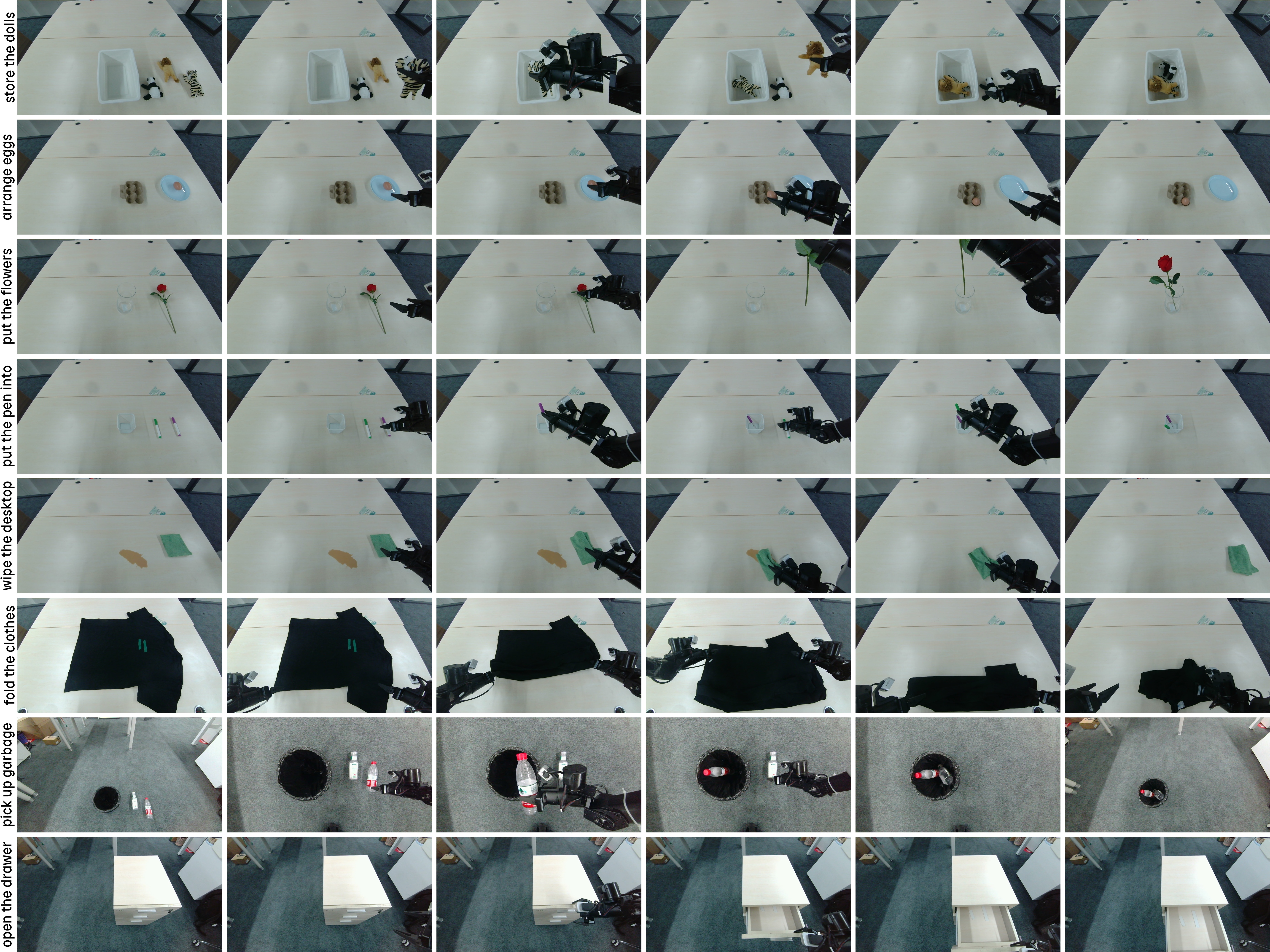}
\caption{\textbf{Cross-scene zero-shot real-robot deployment.} We deploy SimVLA without any additional fine-tuning on held-out scenes and evaluate it on multi-stage tasks (e.g., store dolls, arrange eggs, fold clothes) that require both dexterous manipulation and semantic understanding.}
\label{fig:real_robot_examples}
\end{figure}

The field of Vision-Language-Action (VLA) learning has advanced rapidly, driven by a wave of architectural innovations. Recent methods have proposed increasingly sophisticated designs, ranging from mechanisms that enrich perception with temporal context, to modules that inject explicit 3D spatial awareness, to high-capacity decoders that model complex action distributions. While these contributions have pushed the boundaries of robot capabilities, they also introduce a significant challenge for the research community: attributing performance gains to specific components. Since architectural changes are frequently introduced alongside confounding variables—such as varying pretraining datasets, differing backbone scales, or ad-hoc optimization schedules—it can be difficult to disentangle the impact of a novel mechanism from the underlying training recipe.

To disentangle these confounders, we introduce \textbf{SimVLA}, a deliberately minimal VLA that serves as a \emph{controlled testbed} rather than an architectural novelty. As shown in Fig.~\ref{fig:simvla_overview}, SimVLA decouples perception from control: a standard pretrained vision-language backbone produces fused representations, and a lightweight action head maps them to continuous actions.
We then standardize training dynamics that are often under-specified across papers, including data shuffling, action-space normalization, and optimization schedules. This modular design enables clean ablations along axes that entangled architectures cannot isolate, such as swapping action heads (flow matching, diffusion, and regression) or VLM backbones without redesigning cross-modal adapters.

Through systematic cross-benchmark ablations on four datasets (LIBERO, CALVIN, WidowX, Google Robot), we quantify a consistent \emph{attribution hierarchy} in VLA performance: training dynamics have the largest observed effect ($\Delta$$\approx$\crhl{54--94}\%), task configuration ranks second ($\Delta$$\approx$11--43\%), and architecture has the smallest residual effect ($\Delta$$\approx$0.6--15.6\%). This finding parallels well-established observations in language modeling, where data scale, data quality, and optimization recipes can outweigh architectural changes~\citep{hoffmann2022training,touvron2023llama}; however, this attribution hierarchy has not been systematically quantified in the VLA setting. These observations together form the paper's central empirical claim: an attribution hierarchy that future VLA research can use to interpret and control its reported gains. 

Despite its simplicity, SimVLA is both effective and efficient. Figure~\ref{fig:real_robot_examples} illustrates SimVLA’s cross-scene zero-shot generalization capability when deployed on the Galaxea R1 Lite; additional details about the robot platform and training data are provided in Section~\ref{sec:robot}. Table~\ref{tab:efficiency} further presents a representative example in which our model outperforms multi-billion-parameter baselines while requiring substantially lower peak training VRAM.
\begin{table}[t]
\centering
\begin{minipage}[t]{0.55\linewidth}
  \centering
  \includegraphics[width=\linewidth]{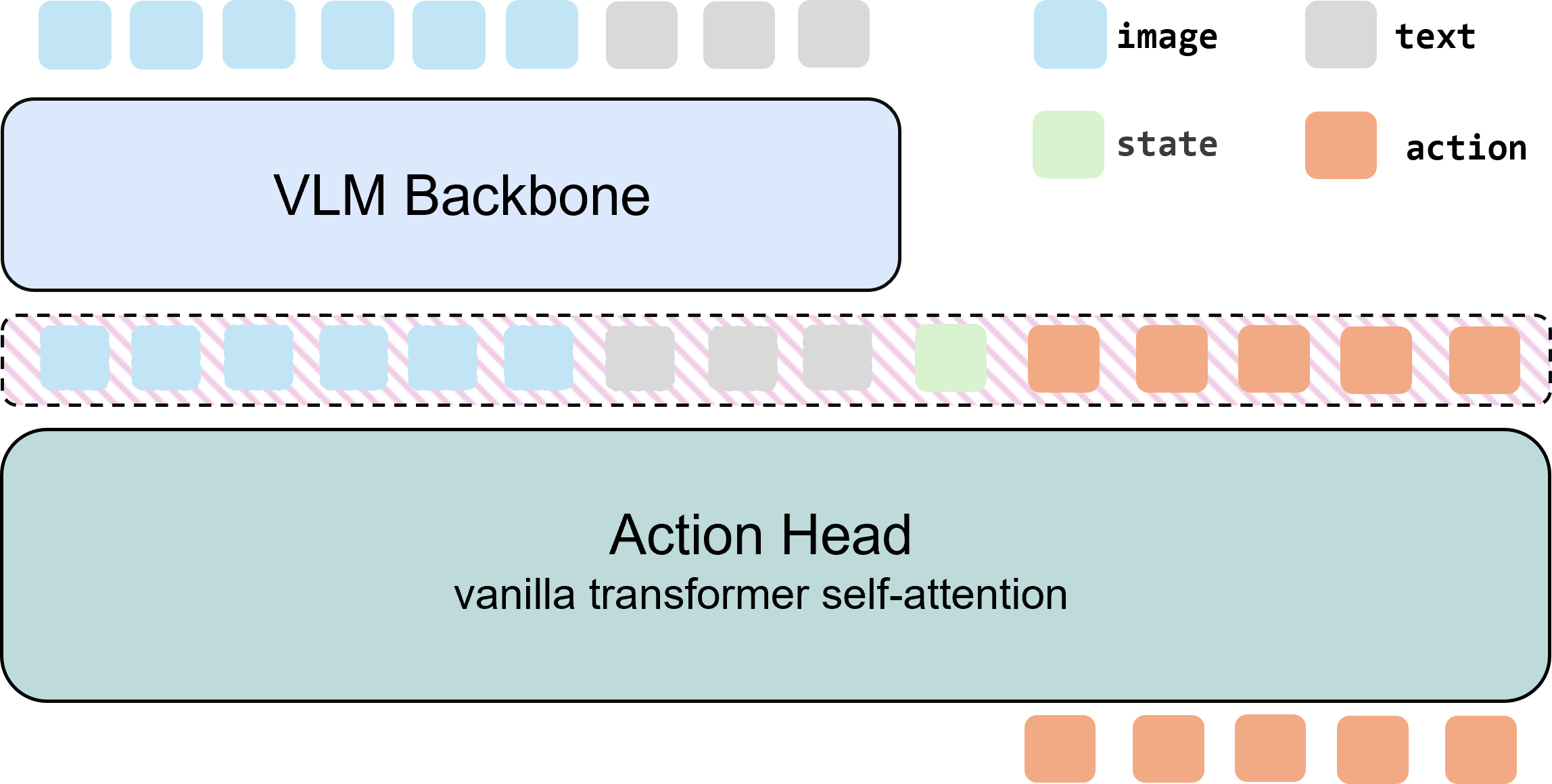}
  \captionof{figure}{\textbf{SimVLA overview.} A pretrained VLM encoder produces fused vision-language tokens once per control step; a lightweight action transformer then outputs a continuous action chunk.}
  \label{fig:simvla_overview}
\end{minipage}%
\hfill%
\begin{minipage}[t]{0.42\linewidth}
  \centering
  
  \vspace{-1.5 in}
  \caption{\textbf{Performance and efficiency summary.} LIBERO success (\%) and peak training VRAM at Batch=8 (GB). ``B'' = backbone scale in billions; SimVLA's 0.5B counts the VLM backbone only. 
  }
  \label{tab:efficiency}
  \scriptsize
  \setlength{\tabcolsep}{3pt}
  \resizebox{\linewidth}{!}{%
  \begin{tabular}{l|c|c|c}
  \toprule
  \textbf{Model} & \textbf{B} & \textbf{LIBERO} & \textbf{VRAM} \\
   & & \textbf{Avg (\%)} & \textbf{(GB)} \\
  \midrule
  OpenVLA-OFT & 7 & 97.1 & 62.0 \\
  $\pi_{0.5}$ & 3 & 96.9 & 51.3 \\
  VLA-Adapter & 0.5 & 97.3 & 24.7 \\
  \rowcolor{gray!10} SimVLA (Ours) & \textbf{0.5} & \textbf{98.6} & \textbf{9.3}  \\
  \bottomrule
  \end{tabular}%
  }
\end{minipage}
\end{table}

\noindent\textbf{Our main contributions are:}
\begin{itemize}[leftmargin=*,noitemsep]
  \item \crhl{\textbf{Methodological and empirical contribution.} We present a controlled attribution study and a minimal-baseline calibration, not novel training techniques. }Through strict one-knob-at-a-time ablations on four benchmarks, we identify a \textbf{quantitative attribution hierarchy} for VLA performance: training dynamics ($\Delta$$\approx$\crhl{54--94}\%) $>$ task configuration ($\Delta$$\approx$11--43\%) $>$ architecture ($\Delta$$\approx$0.6--15.6\%). This hierarchy provides a calibration scale for evaluating future architectural claims under matched training and task-configuration controls\crhl{, within the VLM-encoder plus lightweight-head family}.
  \item We demonstrate that SimVLA—with a small backbone and \textbf{no additional large-scale robot-data VLA pretraining before benchmark-specific finetuning}—achieves state-of-the-art or competitive results on LIBERO (98.6\%), CALVIN ABC$\rightarrow$D (4.47), WidowX (95.8\%), and Google Robot (77.0\%). 
\end{itemize}

\section{Related Work}
\label{sec:related_work}

Recent VLA research has improved robot control through increasingly rich multimodal architectures. We group these advances along three axes and refer interested readers to recent surveys for broader coverage of the field~\citep{VLA_anatomy_survey,VLA_model_survey,GAI_survey}.

\textbf{Visual and Temporal Augmentation.}
Beyond static RGB inputs used by early VLAs such as OpenVLA~\citep{kim2024openvla}, recent methods add explicit motion, prediction, or memory signals. FlowVLA, CoT-VLA, TraceVLA, and 4D-VLA inject optical flow, sub-goal images, trajectory traces, or spatiotemporal features~\citep{FlowVLA,CoT_VLA,TraceVLA,p4D_VLA}; PixelVLA adds pixel-level perception via multiscale encoders~\citep{PixelVLA} and ReconVLA adds a reconstruction objective~\citep{ReconVLA}; WorldVLA, Dream-VLA, ThinkAct, IntentionVLA, FPC-VLA, MemoryVLA, and HAMLET add future prediction, planning, intention reasoning, failure correction, or history-aware memory~\citep{WorldVLA,Dream_VLA,ThinkAct,IntentionVLA,FPC_VLA,MemoryVLA,HAMLET}. These additions improve temporal reasoning but often increase computational overhead and deployment complexity.

\textbf{Geometric and 3D Priors.}
Another line injects spatial structure that 2D VLM backbones may not represent precisely. 4D-VLA and SpatialVLA add 3D positional encodings~\citep{p4D_VLA,SpatialVLA}; GraspVLA and MolmoAct introduce progressive grasp reasoning and visual reasoning traces~\citep{GraspVLA,MolmoAct}; GeoVLA, FALCON, and DepthVLA process point clouds, spatial tokens, or predicted depth features~\citep{GeoVLA,FALCON,DepthVLA}. These designs can improve geometric precision, but they also introduce sensor dependencies or heavy 3D encoders that may reduce embodiment generality.

\textbf{Action Representations and Architectures.}
Recent work also studies how actions should be represented and decoded. FAST, PD-VLA, OpenVLA-OFT, VLA-Adapter, FLOWER, X-VLA, NORA, SmolVLA, GR00T-N1, and UnifiedVLA improve efficiency through tokenization, direct regression, adapters, soft prompts, embodiment-specific designs, or unified autoregressive token modeling~\citep{FAST,PD_VLA,OpenVLA_OFT,VLA_Adapter,FLOWER,x_VLA,NORA,SmolVLA,bjorck2025gr00t,UnifiedVLA}. Diffusion and latent-action policies such as Diffusion Policy, \(\pi_0\), \(\pi_{0.5}\), DD-VLA, UniVLA, UniAct, and VQ-VLA improve action expressivity~\citep{DiffusionPolicy,pi0,pi05,DD_VLA,UniVLA,UniAct,VQ_VLA}, sometimes at the cost of latency or training complexity.

Our work offers a counterpoint to this trend. SimVLA asks whether a simple VLM encoder plus lightweight continuous-action head, paired with a carefully controlled training recipe, can match more elaborate systems. Concurrent VLANeXt~\citep{wu2026vlanext} also argues that VLA progress depends on design and training recipes, and progressively builds a strong model on LIBERO/LIBERO-plus by distilling practical design findings. SimVLA is complementary: rather than optimizing a final recipe, we use a deliberately minimal VLM-encoder-plus-action-head testbed and one-knob-at-a-time ablations across four benchmarks to quantify which reported gains remain after training dynamics and task configuration are controlled. We provide a summary mapping of these aforementioned baselines to their respective design axes in Table~\ref{tab:baseline_categorization} (Appendix).

\section{SimVLA: A Simple VLA Baseline} 
\label{sec:method}

As introduced in \Cref{sec:related_work}, recent VLA models often conflate architecture, task configuration, and training recipe changes, making it difficult to identify which choices actually drive performance. We therefore use SimVLA as a deliberately minimal and standardized empirical testbed: a pretrained VLM encoder coupled to a lightweight continuous-action head. This design fixes the architectural surface enough to support one-knob-at-a-time comparisons, allowing the evaluation in \Cref{sec:evaluation} to estimate an attribution hierarchy over training dynamics, task configuration, and architecture.

\subsection{Preliminaries}
\label{sec:preliminaries}
\textbf{Problem Formulation.} We model the conditional distribution of a future action chunk
\(A_t = [a_t, a_{t+1}, \dots, a_{t+H-1}] \in \mathbb{R}^{H\times d_a}\)
given an observation \(o_t\).
The observation contains the respective multi-view RGB images \(I^1_t,\dots,I^n_t\), the pairing language instruction \(\ell_t\), and the robot proprioception (state) \(s_t\):
\(o_t = [I^1_t,\dots,I^n_t, \ell_t, s_t]\).

\textbf{VLM Backbone Encoder.}
Following the standard late-fusion paradigm \citep{pi0}, we employ a pretrained VLM $E_\phi$ to map multi-view RGB observations and the corresponding language instruction into a shared token representation:
\[
Z_t = E_\phi(I^1_t, \dots, I^n_t, \ell_t).
\]

We treat the VLM as a \emph{perception-language encoder}, not an action generator: semantic understanding is handled by the VLM, while continuous control is delegated to a lightweight action head. The VLM is not frozen by default; we jointly fine-tune it with the action head, optionally using a short warm-up for stability.

\textbf{Action Head.}
SimVLA uses a vanilla Transformer encoder \citep{vaswani2017attention} as a lightweight action head over fused VLM tokens \(Z_t\), proprioception \(s_t\), and action-query tokens. This interface is objective-agnostic: different continuous-action objectives can be instantiated while keeping the VLM backbone, action-head architecture, and training recipe fixed.

\textbf{Default Action Objective.}
The reported SimVLA default instantiates this interface with conditional flow matching \citep{lipman2022flow,pi0}. Let $x$ denote a normalized action chunk and $\epsilon \sim \mathcal{N}(0, I)$ denote Gaussian noise. We sample \(t \in (0,1]\), construct \(x_t=t\epsilon+(1-t)x\), and train the action head \(v_\theta(x_t,o_t,t)\) with:
\[
\mathcal{L}(\theta) = \mathbb{E}\left[\|v_\theta(x_t, o_t, t) - (\epsilon - x)\|_2^2\right].
\]

At inference time, the default head integrates the learned vector field with a small number of Euler steps to generate smooth action chunks. Under identical training conditions, flow matching outperforms DDPM diffusion and MSE regression in our LIBERO action-objective ablation (98.6\% vs.\ 95.5\% and 95.3\%; Table~\ref{tab:libero_ablations}). The default training and inference settings employed by SimVLA are detailed in the Appendix (Table~\ref{tab:hparams_libero}).

\subsection{SimVLA Architecture}
\label{sec:architecture}
\textbf{Design Principle.}
SimVLA adopts an intentionally minimal design (Figure~\ref{fig:simvla_overview}): a vision-language encoder produces fused representations once per control step, and a lightweight action transformer generates continuous action chunks under a configurable action objective.

At each timestep, the fused vision-language tokens $Z_t$ are first obtained from the VLM backbone. We then construct the input sequence to the action head by concatenating projected VLM tokens, a broadcasted proprioceptive state embedding, a timestep embedding, and objective-specific action-query tokens. This unified token sequence is processed by a vanilla Transformer encoder with pure self-attention \citep{vaswani2017attention}, without cross-attention, memory modules, or modality-specific routing.

\textbf{Rationale and Practical Advantages.}
Self-attention over concatenated tokens provides a neutral integration mechanism: more specialized bridges or conditional normalization may help, but they also add confounders. Executing the VLM once per control step and delegating action generation to the lightweight head also reduces inference cost.

\subsection{Controlled Training and Inference Setup}
\label{sec:training}
For SimVLA to serve as an attribution testbed, its reference setup must be specified as part of the experimental control rather than treated as incidental implementation detail. We therefore define the action representation, data handling, optimization, architecture configuration, and inference procedure used by the policy; ablations then vary one of these choices while holding the others fixed.

\textbf{Action Representation and Normalization.}
The policy predicts normalized continuous action chunks of horizon $H$ and executes them in a receding-horizon manner. Normalization statistics are computed per action dimension from the training set, and proprioceptive states are normalized when applicable. We set \(H{=}10\) for LIBERO and \(H{=}30\) for CALVIN, WidowX, Google Robot, and real-robot training, with horizon changes treated as controlled task-configuration ablations.

\textbf{Data Handling.}
Training uses trajectory shuffling as the reference data-ordering choice. Because demonstration ordering can affect optimization and long-horizon generalization, we treat shuffling as an explicit controlled variable rather than as a hidden data-loader choice.

\textbf{Optimization Dynamics.}
For optimization, we fix batch size, training budget, learning-rate schedule, and backbone adaptation settings for each benchmark. In the main LIBERO configuration, this corresponds to AdamW with learning rate \(2\!\times\!10^{-4}\), no warm-up, no scheduler, and a VLM learning-rate multiplier of 0.1. Ablations vary these optimization knobs while holding the architecture and task configuration fixed.

\textbf{Architecture Configuration.}
We treat action transformer scale, VLM backbone choice, and information injection (concatenation, cross-attention, AdaLN) as controlled architecture knobs evaluated only after data and optimization controls are matched. The reference policy uses the large action head \(\{1024,24,16\}\) for \{hidden size, depth, heads\} \crhl{($\sim$300M parameters, $\sim$0.80B trainable parameters in total with the 0.5B backbone)}\crhl{; the small head \(\{768,12,12\}\) has $\sim$80M parameters ($\sim$0.58B in total). Throughout the paper, ``B'' denotes backbone scale only.} Benchmark-specific backbone choices, optimizer settings, and budgets are in the Appendix.

\textbf{Inference.}
At inference time, SimVLA follows an \emph{encode-once, predict-in-the-head} workflow. Given an observation, the VLM backbone is executed once to obtain fused tokens, after which the lightweight action head generates a continuous action chunk under the configured objective. The resulting actions are post-processed and executed in a receding-horizon fashion.

\section{Evaluation}
\label{sec:evaluation}

\subsection{Experimental Setup}

We follow standard evaluation protocols for each benchmark.

\textbf{LIBERO \citep{liu2023libero}.} We utilize all four standard suites: LIBERO-Spatial, LIBERO-Object, LIBERO-Goal, and LIBERO-Long, comprising 10 tasks per suite with 500 expert demonstrations each. This serves as our primary testbed to evaluate the model's long-horizon consistency and generalization performance.

\textbf{LIBERO-PRO \citep{zhou2025libero}.} To address the memorization issue in current VLA studies, we evaluate on LIBERO-PRO, a robust benchmark that introduces systematic perturbations across four dimensions: object appearance (Obj), spatial layout (Pos), language instructions (Sem), and task goals (Task). 


\textbf{SimplerEnv \citep{li2024evaluating}.} We evaluate on Simpler-Fractal (Google Robot) and Simpler-Bridge (WidowX) to assess the model's performance in high-fidelity simulated environments. For Simpler-Fractal, we report variant aggregation scores to test the policy's robustness against diverse scene variations. For Simpler-Bridge, the evaluation focuses on real-to-sim transfer across tasks.

\textbf{CALVIN \citep{mees2022calvin}.} We use the standard ABC$\rightarrow$D setting to evaluate long-horizon language-conditioned manipulation and cross-scene zero-shot generalization to an unseen environment. Following the official protocol, each evaluation episode contains a chain of five language instructions, and we report the success rates of completing at least 1/2/3/4/5 consecutive subtasks, together with the average successful sequence length.

\textbf{Evaluation Protocol.} Following recent works \citep{x_VLA,DD_VLA,VLA_Adapter}, we train a \emph{single generalist policy} on the union of all standard LIBERO datasets (Spatial, Object, Goal, Long). For LIBERO-PRO, we directly evaluate this policy without additional fine-tuning to strictly test cross-scene zero-shot robustness.
For SimplerEnv, we train our model on the Fractal~\citep{brohan2022rt} and BridgeData-V2 datasets~\citep{walke2023bridgedata}, strictly following recent work~\citep{x_VLA}. For CALVIN, we train on the standard ABC split and evaluate cross-scene zero-shot on the held-out D environment under the official multi-step evaluation protocol~\citep{x_VLA}.
Across all benchmarks, we train directly from a pretrained VLM backbone without additional large-scale robot-data VLA pretraining before benchmark-specific training.

\textbf{Critical Hyperparameters.} A key finding of this study is that performance is highly sensitive to a small set of training and design choices, which can easily overshadow architectural differences if left under-tuned. To make our comparisons transparent, we summarize the concrete settings we tried:
\begin{itemize}[leftmargin=*,noitemsep,topsep=0pt]

    \item \textbf{Data \& representation.}
    We vary the \textit{action-chunk horizon} \(H\in\{10,20,30\}\).
    We ablate \textit{data shuffling} (shuffle vs.\ no shuffle).
    We ablate \textit{normalization} (on/off) for both actions and proprioception; when enabled, we compute per-dimension statistics (mean/std, with robust quantile estimates) on the training set and normalize accordingly.

    \item \textbf{Optimization dynamics.}
    We sweep the \textit{learning rate} over \(\{5\!\times\!10^{-5}, 10^{-4}, 2\!\times\!10^{-4}, 5\!\times\!10^{-4}\}\),
    \textit{warm-up steps} over \(\{0,1000\}\), and \textit{the scheduler} over \{cosine decay, none\}.
    For all variants, we keep the training budget fixed (batch size and total steps/epochs).
    
    \item \textbf{Architecture configuration.}
    \textit{Action transformer scale:} we use two configurations (small vs.\ large), as in our training scripts, e.g., \(\{768,12,12\}\) (\(\sim\)80M params) vs.\ \(\{1024,24,16\}\) (\(\sim\)300M params) for \{hidden size, depth, \# heads\}.
    \textit{VLM backbone:} we compare Florence-2 (0.9B) \citep{xiao2024florence} and SmolVLM-0.5B \citep{marafioti2025smolvlm}\crhl{, and additionally evaluate Qwen3-VL-2B \citep{qwen3vl} at the default LIBERO configuration (Appendix~\ref{app:robust_backbone})}.
    \textit{Information injection:} we compare (i) token concatenation with pure self-attention (default), (ii) cross-attention injection, and (iii) conditional AdaLN.
\end{itemize}

Detailed hyperparameters are provided in the Appendix~\ref{app:training_hparams}. 

\textbf{Baselines.} We compare against 34 representative VLA policies \crhl{using the results reported in their original papers (none are reproduced under our setup). SimVLA numbers in the tables are single seeds (seed~0); three-seed results are in Appendix~\ref{app:robust_seeds} (Tables~\ref{tab:app_seed_libero}--\ref{tab:app_seed_calvin}).}

\begin{table}[t]
\centering
\begin{minipage}[t]{0.50\linewidth}
  \centering
  \caption{\textbf{Comparison on the LIBERO benchmark.} Success rate (\%) per suite and overall average. \textbf{Bold$^*$} = best; \textbf{Bold} = second-best. ``B'' = backbone scale in billions.}
  \label{tab:libero_comparison}
  \scriptsize
  \setlength{\tabcolsep}{2pt}
  \resizebox{\linewidth}{!}{%
  \begin{tabular}{l|c|cccc|c}
  \toprule
  \textbf{Model} & \textbf{B} & \textbf{Spatial} & \textbf{Object} & \textbf{Goal} & \textbf{Long} & \textbf{Avg} \\
  \midrule
  \multicolumn{7}{l}{\textit{\textbf{Large Models ($\ge$ 4B)}}} \\
  \midrule
  UniVLA \citep{UniVLA} & 9 & 96.5 & 96.8 & 95.6 & 92.0 & 95.2 \\
  FlowVLA \citep{FlowVLA} & 8.5 & 93.2 & 95.0 & 91.6 & 72.6 & 88.1 \\
  UnifiedVLA \citep{UnifiedVLA} & 8.5 & 95.4 & 98.8 & 93.6 & 94.0 & 95.5 \\
  OpenVLA \citep{kim2024openvla} & 7 & 84.7 & 88.4 & 79.2 & 53.7 & 76.5 \\
  OpenVLA-OFT \citep{OpenVLA_OFT} & 7 & 97.6 & 98.4 & 97.9 & 94.5 & 97.1 \\
  DD-VLA \citep{DD_VLA} & 7 & 97.2 & 98.6 & 97.4 & 92.0 & 96.3 \\
  MemoryVLA \citep{MemoryVLA} & 7 & 98.4 & 98.4 & 96.4 & 93.4 & 96.7 \\
  PD-VLA \citep{PD_VLA} & 7 & 95.5 & 96.7 & 94.9 & 91.7 & 94.7 \\
  MolmoAct \citep{MolmoAct} & 7 & 87.0 & 95.4 & 87.6 & 77.2 & 86.6 \\
  ThinkAct \citep{ThinkAct} & 7 & 88.3 & 91.4 & 87.1 & 70.9 & 84.4 \\
  CoT-VLA \citep{CoT_VLA} & 7 & 87.5 & 91.6 & 87.6 & 69.0 & \crhl{83.9} \\
  WorldVLA \citep{WorldVLA} & 7 & 87.6 & 96.2 & 83.4 & 60.0 & 81.8 \\
  TraceVLA \citep{TraceVLA} & 7 & 84.6 & 85.2 & 75.1 & 54.1 & 74.8 \\
  FPC-VLA \citep{FPC_VLA} & 7 & \crhl{87.0} & \crhl{92.0} & \crhl{86.2} & 82.2 & 86.9 \\
  4D-VLA \citep{p4D_VLA} & 4 & 88.9 & 95.2 & 90.9 & 79.1 & 88.6 \\
  SpatialVLA \citep{SpatialVLA} & 4 & 88.2 & 89.9 & 78.6 & 55.5 & 78.1 \\
  \midrule
  \multicolumn{7}{l}{\textit{\textbf{Small Models ($<$ 4B)}}} \\
  \midrule
  $\pi_0$ \citep{pi0} & 3 & 96.8 & 98.8 & 95.8 & 85.2 & 94.2 \\
  $\pi_0$-FAST \citep{FAST} & 3 & 96.4 & 96.8 & 88.6 & 60.2 & 85.5 \\
  $\pi_{0.5}$ \citep{pi05} & 3 & 98.8 & 98.2 & {\textbf{98.0}} & 92.4 & 96.9 \\
  NORA \citep{NORA} & 3 & 92.2 & 95.4 & 89.4 & 74.6 & 87.9 \\
  SmolVLA \citep{SmolVLA} & 2.2 & 93.0 & 94.0 & 91.0 & 77.0 & 88.8 \\
  VLANeXt \citep{wu2026vlanext} & 2 & {\textbf{99.0}} & 99.2 & 96.6 & 94.6 & 97.4 \\
  GR00T-N1 \citep{bjorck2025gr00t} & 2 & 94.4 & 97.6 & 93.0 & 90.6 & 93.9 \\
  GraspVLA \citep{GraspVLA} & 1.8 & - & 94.1 & 91.2 & 82.0 & 89.1 \\
  FLOWER \citep{FLOWER} & 1 & \crhl{97.5} & \crhl{99.1} & \crhl{96.1} & \crhl{94.9} & \crhl{96.9} \\
  \midrule
  \multicolumn{7}{l}{\textit{\textbf{Tiny Models ($<$ 1B)}}} \\
  \midrule
  X-VLA \citep{x_VLA} & 0.9 & 98.2 & 98.6 & 97.8 & \textbf{97.6$^*$} & {\textbf{98.1}} \\
  VLA-Adapter \citep{VLA_Adapter} & 0.5 & 97.8 & {\textbf{99.2}} & 97.2 & 95.0 & 97.3 \\
  VLA-OS \citep{VLA_OS} & 0.5 & 87.0 & 96.5 & 92.7 & 66.0 & 85.6 \\
  UniAct \citep{UniAct} & 0.5 & 77.0 & 87.0 & 77.0 & 70.0 & 77.8 \\
  \midrule
  \rowcolor{gray!10} \textbf{SimVLA} & 0.5 & \textbf{99.6$^*$} & \textbf{99.8$^*$} & \textbf{98.6$^*$} & \textbf{96.4} & \textbf{98.6$^*$} \\
  \bottomrule
  \end{tabular}%
  }
\end{minipage}%
\hfill%
\begin{minipage}[t]{0.48\linewidth}
  \centering
  \captionof{table}{\textbf{Robustness on LIBERO-PRO.} Success rate (\%) across perturbation dimensions: Original (Ori), Object (Obj), Position (Pos), Semantic (Sem), Task (Task). \textbf{Bold} = best.}
  \label{tab:libero_pro_full}
  \scriptsize
  \setlength{\tabcolsep}{2pt}
  \resizebox{\linewidth}{!}{%
  \begin{tabular}{l|l|ccccc}
  \toprule
  \textbf{Suite} & \textbf{Method} & \textbf{Ori} & \textbf{Obj} & \textbf{Pos} & \textbf{Sem} & \textbf{Task} \\
  \midrule
  \multirow{3}{*}{Spatial}
  & OpenVLA \citep{kim2024openvla} & 98.0 & 97.0 & 0.0 & 97.0 & 0.0 \\
  & $\pi_{0.5}$ \citep{pi05} & 98.0 & 97.0 & 20.0 & 97.0 & \textbf{1.0} \\
  \rowcolor{gray!10} & \textbf{SimVLA} & \textbf{99.0} & \textbf{98.0} & \textbf{29.0} & \textbf{98.0} & 0.0 \\
  \midrule
  \multirow{3}{*}{Object}
  & OpenVLA \citep{kim2024openvla} & 99.0 & \textbf{98.0} & 0.0 & 98.0 & 0.0 \\
  & $\pi_{0.5}$ \citep{pi05} & 98.0 & \textbf{98.0} & \textbf{17.0} & 96.0 & 1.0 \\
  \rowcolor{gray!10} & \textbf{SimVLA} & \textbf{100.0} & 85.0 & 1.0 & \textbf{100.0} & \textbf{4.0} \\
  \midrule
  \multirow{3}{*}{Goal}
  & OpenVLA \citep{kim2024openvla} & 98.0 & 96.0 & 0.0 & 98.0 & 0.0 \\
  & $\pi_{0.5}$ \citep{pi05} & 97.0 & \textbf{97.0} & \textbf{38.0} & 97.0 & 0.0 \\
  \rowcolor{gray!10} & \textbf{SimVLA} & \textbf{99.0} & 82.0 & 0.0 & \textbf{99.0} & \textbf{10.0} \\
  \midrule
  \multirow{3}{*}{Long}
  & OpenVLA \citep{kim2024openvla} & 93.0 & 81.0 & 0.0 & 96.0 & 0.0 \\
  & $\pi_{0.5}$ \citep{pi05} & 93.0 & \textbf{92.0} & \textbf{8.0} & 93.0 & 1.0 \\
  \rowcolor{gray!10} & \textbf{SimVLA} & \textbf{96.0} & 61.0 & 3.0 & \textbf{98.0} & \textbf{10.0} \\
  \bottomrule
  \end{tabular}%
  }
  \vspace{1.6em}
  \captionof{table}{\textbf{Comparison on CALVIN ABC$\rightarrow$D;} success rate (\%).}
  \label{tab:calvin_results}
  \scriptsize
  \setlength{\tabcolsep}{3pt}
  \resizebox{\linewidth}{!}{%
  \begin{tabular}{l|ccccc|c}
  \toprule
  \textbf{Model} & \textbf{SR1} & \textbf{SR2} & \textbf{SR3} & \textbf{SR4} & \textbf{SR5} & \textbf{Avg.len} \\
  \midrule
  \multicolumn{7}{l}{\textit{\textbf{Large ($\ge$ 4B)}}} \\
  \midrule
  UniVLA \citep{UniVLA} & 95.5 & 85.8 & 75.4 & 66.9 & 56.5 & 3.80 \\
  UnifiedVLA \citep{UnifiedVLA} & 98.9 & 94.8 & 89.0 & 82.8 & 75.1 & 4.41 \\
  OpenVLA \citep{kim2024openvla} & 91.3 & 77.8 & 62.0 & 52.1 & 43.5 & 3.27 \\
  OpenVLA-OFT \citep{OpenVLA_OFT} & 96.3 & 89.1 & 82.4 & 75.8 & 66.5 & 4.10 \\
  HiF-VLA \citep{HiF-VLA} & 98.5 & 94.1 & 88.1 & 81.4 & 73.1 & 4.35 \\
  \midrule
  \multicolumn{7}{l}{\textit{\textbf{Tiny ($<$ 1B)}}} \\
  \midrule
  Seer \citep{Seer} & 96.3 & 91.6 & 86.1 & 80.3 & 74.0 & 4.28 \\
  VLA-Adapter \citep{VLA_Adapter} & 99.1 & 94.6 & 88.8 & 82.8 & 76.5 & 4.42 \\
  X-VLA \citep{x_VLA} & 97.1 & 92.6 & 88.5 & \textbf{84.4} & \textbf{78.8} & 4.43 \\
  \midrule
  \rowcolor{gray!15} \textbf{SimVLA} & \textbf{99.4} & \textbf{96.4} & \textbf{89.8} & 83.7 & 77.2 & \textbf{4.47} \\
  \bottomrule
  \end{tabular}%
  }
\end{minipage}
\end{table}

\begin{table}[t]
\centering
\begin{minipage}[t]{0.50\linewidth}
  \centering
  \caption{\textbf{Comparison on WidowX robot tasks;} success rates (\%).}
  \label{tab:widowx_robot_eval}
  \scriptsize
  \setlength{\tabcolsep}{2pt}
  \resizebox{\linewidth}{!}{%
  \begin{tabular}{l|cccc|c}
  \toprule
  \textbf{Method} & \textbf{Spoon} & \textbf{Carrot} & \textbf{Stack} & \textbf{Eggplant} & \textbf{Avg} \\
  \midrule
  \multicolumn{6}{l}{\textit{\textbf{Large ($\ge$ 4B)}}} \\ \midrule
  RT-1-X \citep{o2024open} & 0.0 & 4.2 & 0.0 & 0.0 & \crhl{1.1} \\
  Octo-Base \citep{Octo} & 12.5 & 8.3 & 0.0 & 43.1 & \crhl{16.0} \\
  OpenVLA \citep{kim2024openvla} & 0.0 & 0.0 & 0.0 & 4.1 & \crhl{1.0} \\
  OpenVLA-OFT \citep{OpenVLA_OFT} & 12.5 & 4.2 & 8.3 & 37.5 & \crhl{15.6} \\
  DD-VLA \citep{DD_VLA} & 29.2 & {29.2} & {20.8} & {70.8} & {\crhl{37.5}} \\
  RoboVLM \citep{liu2025towards} & 29.2 & 25.0 & 12.5 & 58.3 & \crhl{31.3} \\
  SpatialVLA \citep{SpatialVLA} & 16.7 & 25.0 & 29.2 & 100.0 & 42.7 \\
  MemoryVLA \citep{MemoryVLA} & 75.0 & 75.0 & 37.5 & 100.0 & 71.9 \\
  ThinkAct \citep{ThinkAct} & 58.3 & 37.5 & 8.7 & 70.8 & 43.8 \\
  FPC-VLA \citep{FPC_VLA} & \crhl{79.2} & \crhl{58.3} & \crhl{45.8} & 75.0 & 64.6 \\
  \midrule
  \multicolumn{6}{l}{\textit{\textbf{Small ($<$ 4B)}}} \\ \midrule
  Octo-Small \citep{Octo} & {47.2} & 9.7 & 4.2 & 56.9 & \crhl{29.5} \\
  $\pi_0$ \citep{pi0} & 29.1 & 0.0 & {16.7} & 62.5 & \crhl{27.1} \\
  $\pi_0$-FAST \citep{FAST} & 29.1 & 21.9 & 10.8 & {66.6} & \crhl{32.1} \\
  GR00T-N1 \citep{bjorck2025gr00t} & 62.5 & 45.8 & {16.7} & 20.8 & {\crhl{36.5}} \\
  FLOWER \citep{FLOWER} & 71.0 & 13.0 & 8.0 & 88.0 & 45.0 \\
  \midrule
  \multicolumn{6}{l}{\textit{\textbf{Tiny ($<$ 1B)}}} \\ \midrule
  X-VLA \citep{x_VLA} & 100 & 91.7 & 95.8 & 95.8 & 95.8 \\
  \rowcolor{gray!15} \textbf{SimVLA} & \textbf{100} & \textbf{100} & \textbf{91.7} & \textbf{91.7} & \textbf{95.8} \\
  \bottomrule
  \end{tabular}%
  }
\end{minipage}%
\hfill%
\begin{minipage}[t]{0.48\linewidth}
  \centering
  \caption{\textbf{Comparison on Google Robot tasks;} success rates (\%). Variant aggregation.}
  \label{tab:google_robot_eval}
  \scriptsize
  \setlength{\tabcolsep}{2pt}
  \resizebox{0.92\linewidth}{!}{%
  \begin{tabular}{l|ccc|c}
  \toprule
  \textbf{Model} & \textbf{Pick} & \textbf{Move} & \textbf{Open} & \textbf{Avg} \\
  \midrule
  \multicolumn{5}{l}{\textit{\textbf{Large ($\ge$ 4B)}}} \\
  \midrule
  RT-1-X \citep{o2024open} & 49.0 & 32.3 & 29.4 & 36.9 \\
  RT-2-X \citep{o2024open} & 82.3 & 79.2 & 35.3 & 65.6 \\
  Octo-Base \citep{Octo} & 0.6 & 3.1 & 1.1 & 1.6 \\
  OpenVLA \citep{kim2024openvla} & 54.5 & 47.7 & 17.7 & 40.0 \\
  OpenVLA-OFT \citep{OpenVLA_OFT} & 65.3 & 59.0 & 12.2 & 45.5 \\
  RoboVLM \citep{liu2025towards} & 75.6 & 60.0 & 10.6 & 48.7 \\
  TraceVLA \citep{TraceVLA} & 60.0 & 56.4 & 31.0 & 49.1 \\
  DD-VLA \citep{DD_VLA} & 82.5 & 64.6 & 23.6 & 56.9 \\
  SpatialVLA \citep{SpatialVLA} & 88.0 & 72.7 & 41.8 & 67.5 \\
  ThinkAct \citep{ThinkAct} & 84.0 & 63.8 & 47.6 & 65.1 \\
  \midrule
  \multicolumn{5}{l}{\textit{\textbf{Small ($<$ 4B)}}} \\
  \midrule
  $\pi_0$ \citep{pi0} & 75.2 & 63.7 & 25.6 & 54.8 \\
  $\pi_0$-FAST \citep{FAST} & 77.6 & 68.2 & 31.3 & 59.0 \\
  GR00T-N1 \citep{bjorck2025gr00t} & 78.8 & 62.5 & 13.2 & 51.5 \\
  \midrule
  \multicolumn{5}{l}{\textit{\textbf{Tiny ($<$ 1B)}}} \\
  \midrule
  X-VLA \citep{x_VLA} & 85.5 & 79.8 & 61.9 & 75.7 \\
  \rowcolor{gray!15} \textbf{SimVLA} & \textbf{82.3} & \textbf{81.0} & \textbf{67.7} & \textbf{77.0} \\
  \bottomrule
  \end{tabular}%
  }
\end{minipage}
\end{table}

\textbf{Real-robot (Galaxea R1 Lite).} \label{sec:robot}
Beyond simulation, we evaluate cross-scene zero-shot generalization on a mobile bimanual robot. We train SimVLA and a $\pi_{0.5}$ baseline on the 500-hour Galaxea Open-World Dataset~\citep{jiang2025galaxea}, then deploy both in held-out scenes without additional fine-tuning. We evaluate on eight multi-stage manipulation tasks that emphasize {dexterous, fine-grained manipulation} under diverse scenes and objects: \emph{store the dolls}, \emph{arrange eggs}, \emph{put the flowers in the vase}, \emph{put the pen into the pen holder}, \emph{wipe the desktop}, \emph{fold the clothes}, \emph{pick up garbage on the ground}, and \emph{open the drawer}. For each task, we report task success under a fixed time budget. Additional dataset/robot details and per-task rubrics are provided in Appendix~\ref{app:galaxea_dataset}.

\subsection{Results}

\textbf{Results on LIBERO.} {Table~\ref{tab:libero_comparison} shows that SimVLA achieves the highest reported average success rate among the compared baselines. Despite using a compact 0.5B backbone, it surpasses large baselines ($\ge$4B) such as OpenVLA-OFT (97.1\%) and MemoryVLA (96.7\%), demonstrating the strength of a simple, well-tuned architecture. SimVLA achieves near-perfect scores on \textit{Spatial} (99.6\%), \textit{Object} (99.8\%), and \textit{Goal} (98.6\%), ranking first overall. On the challenging \textit{Long} suite, it attains 96.4\%, demonstrating strong temporal consistency without explicit memory modules. 

\textbf{Robustness on LIBERO-PRO.} Table~\ref{tab:libero_pro_full} presents our evaluation on the rigorous LIBERO-PRO benchmark. SimVLA demonstrates strong generalization, particularly under \textbf{Semantic} and \textbf{Task} perturbations. While OpenVLA and $\pi_{0.5}$ drop to near-zero performance on task-level changes, SimVLA retains nonzero generalization, reaching \textbf{10.0\%} success on both \textit{Goal} and \textit{Long} suites. SimVLA also ranks first across all suites in Semantic robustness with $\sim$99\% success and performs best on Spatial Position robustness (29.0\% in \textit{Spatial}). \crhl{SimVLA is, however, clearly weaker under Object and Position perturbations on the long-horizon suites (Object-Long 61.0\%, Position-Long 3.0\%). We do not attribute this gap to a single component, since $\pi_{0.5}$ differs from SimVLA simultaneously in backbone scale, robot-data VLA pretraining, architecture, and training recipe. Disentangling these factors is future work.}

\textbf{Results on WidowX.} As detailed in Table~\ref{tab:widowx_robot_eval}, SimVLA achieves a 95.8\% overall average, matching the heavily pre-trained X-VLA while using only benchmark-specific training from a pretrained VLM initialization. It also secures perfect scores (100\%) on two displayed tasks and substantially exceeds large-scale baselines such as MemoryVLA (71.9\%) and FPC-VLA (64.6\%).

\textbf{Results on Google Robot.} In the Google Robot variant aggregation tasks shown in Table~\ref{tab:google_robot_eval}, SimVLA achieves 77.0\% average success, outperforming strong baselines such as SpatialVLA (67.5\%), RT-2-X (65.6\%), and ThinkAct (65.1\%). SimVLA is also slightly higher than X-VLA (75.7\%) on the overall average, although the margin is small. Together, these results suggest that a simple, data-efficient baseline can remain competitive on challenging benchmarks without extensive VLA pretraining.

\textbf{Results on CALVIN.} Table~\ref{tab:calvin_results} shows that SimVLA achieves the best overall average successful sequence length of 4.47 on CALVIN ABC$\rightarrow$D, slightly above compact baselines such as VLA-Adapter (4.42) and X-VLA (4.43), while also surpassing larger models including OpenVLA-OFT (4.10) and UniVLA (3.80). Despite using only a small backbone, SimVLA reaches 99.4\% / 96.4\% / 89.8\% success on completing 1/2/3 consecutive subtasks, indicating strong long-horizon consistency in the unseen D environment.

\textbf{Open-world evaluation (BEHAVIOR-1K).} As preliminary evidence beyond fixed-task benchmarks, we evaluate SimVLA on the BEHAVIOR-1K Challenge~\citep{BEHAVIOR1K}. On the scored Task~1 (\emph{Turning on Radio}), SimVLA achieves a task success score of 0.40, outperforming $\pi_{0.5}$ and approaching the 1st-place challenge winner~\citep{larchenko2025behavior}. Full details are in the Appendix~\ref{app:behavior1k}.

\subsection{Ablation Studies}
\label{sec:ablations}

\textbf{Cross-benchmark attribution hierarchy.}
We conduct one-knob-at-a-time ablations, varying each factor while holding all other settings fixed. Table~\ref{tab:libero_ablations} reports the main LIBERO audit; Appendix Tables~\ref{tab:calvin_ablations}--\ref{tab:simpler_ablations} extend the same protocol to CALVIN, WidowX, and Google Robot. Across benchmarks, training dynamics (data shuffling, action normalization, VLM LR multiplier) produce the largest swings ($\Delta$$\approx$\crhl{54--94}\%), followed by task configuration ($\Delta$$\approx$11--43\%) and architecture ($\Delta$$\approx$0.6--15.6\%). These deltas are measured relative to our default configuration and do not exhaustively characterize knob interactions, but the consistent ordering is the paper's central empirical finding.

\crhl{Appendix~\ref{app:robustness} repeats the sweeps around alternative base configurations (a small action head, a Florence-2 backbone, and an autoregressive action head) and reports three-seed results for the default SimVLA and for Table~\ref{tab:libero_ablations}; the overall ordering is unchanged.}

\begin{table}[t]
\caption{\textbf{Ablations on LIBERO.} Each row corresponds to one ablation setting with the remaining knobs fixed to the default configuration. The default row matches the SimVLA entry in Table~\ref{tab:libero_comparison} and serves as the reference point for the controlled one-knob-at-a-time comparisons. \crhl{Three-seed std.: Appendix Table~\ref{tab:app_libero_3seed}.}}
\label{tab:libero_ablations}
\centering
\scriptsize
\setlength{\tabcolsep}{3pt}
\resizebox{0.88\textwidth}{!}{%
\begin{tabular}{l|l|ccccc}
\toprule
\textbf{Knob} & \textbf{Value} & \textbf{Spatial} & \textbf{Object} & \textbf{Goal} & \textbf{Long} & \textbf{Avg} \\
\midrule
\rowcolor{gray!10} \textbf{SimVLA} & Default settings & \textbf{99.6} & \textbf{99.8} & \textbf{98.6} & \textbf{96.4} & \textbf{98.6} \\
\midrule
\rowcolor{gray!10} \multicolumn{7}{l}{\textit{Data \& representation} \ (default: \(H{=}10\), shuffling on, normalization on)} \\
\multirow{2}{*}{Action chunk horizon \(H\)} & \(H=20\) & 99.2 & 89.6 & 92.4 & 88.4 & 92.4 \\
& \(H=30\) & 95.4 & 93.8 & 80.6 & 79.2 & 87.3 \\
\midrule
Data shuffling & off & 6.2 & 0.0 & 13.6 & 0.0 & \crhl{5.0} \\
\midrule
Action normalization & off & 22.6 & 3.2 & 23.2 & 0.0 & 12.3 \\
\midrule
\rowcolor{gray!10} \multicolumn{7}{l}{\textit{Optimization dynamics} \ (default: lr \(2\!\times\!10^{-4}\), warm-up none, scheduler none, VLM LR multiplier 0.1)} \\
\multirow{3}{*}{Learning rate}
& \(5\!\times\!10^{-5}\) & 98.0 & 97.6 & 96.2 & 70.4 & 90.6 \\
& \(1\!\times\!10^{-4}\) & 99.6 & 98.2 & 98.4 & 85.6 & 95.5 \\
& \(5\!\times\!10^{-4}\) & 84.4 & 91.8 & 76.0 & 38.4 & 72.7 \\
\midrule
Warm-up steps & 1000 & 97.4 & 99.6 & 96.4 & 93.8 & 96.8 \\
\midrule
Scheduler & cosine & 99.2 & 99.4 & 98.4 & 93.0 & 97.5 \\
\midrule
VLM LR multiplier & 1.0 & 41.2 & 80.8 & 46.4 & 8.2 & 44.2 \\
\midrule
\rowcolor{gray!10} \multicolumn{7}{l}{\textit{Architecture configuration} \ (default: large head (1024,24,16), concat injection, SmolVLM-0.5B, flow matching)} \\
Action transformer scale & small (768,12,12) & 98.8 & 99.6 & 98.6 & 94.8 & 98.0 \\
\midrule
\multirow{2}{*}{Condition injection} & conditional AdaLN & 99.2 & 98.0 & 96.6 & 70.4 & 91.1 \\
& cross-attention & 98.4 & 96.6 & 94.8 & 76.2 & 91.5 \\
\midrule
VLM backbone & Florence-2 & 99.8 & 99.2 & 98.0 & 93.8 & 97.7 \\
\midrule
\multirow{2}{*}{Action objective} & DDPM diffusion & 96.6 & 99.4 & 96.2 & 89.8 & 95.5 \\
& MSE regression & 97.2 & 96.2 & 96.4 & 91.2 & 95.3 \\
\bottomrule
\end{tabular}%
}
\end{table}


\textbf{Attribution findings and control implications.}
The hierarchy implies that VLA comparisons should first match the data pipeline (shuffling and per-dimension normalization), optimizer/backbone adaptation (LR and VLM LR multiplier), and action horizon before attributing gains to architecture. In practice, we recommend prioritizing strong shuffling and normalization, using an LR around $1\!\times\!10^{-4}$ with a reduced VLM LR multiplier, matching $H$ to task complexity, and then comparing action objectives or architectural variants against the matched SimVLA baseline.

\textbf{Results on Real Robot.}
On the eight Galaxea R1 Lite tasks, SimVLA is broadly comparable to $\pi_{0.5}$ under the same cross-scene zero-shot protocol (Fig.~\ref{fig:galaxea_zero_shot}). Most tasks reach around 80\% success, while \emph{fold the clothes}, \emph{put the pen into the pen holder}, and \emph{put the flowers in the vase} remain challenging. SimVLA is trained directly from a pretrained VLM, whereas $\pi_{0.5}$ uses its released VLA initialization.

\begin{figure}[t]
\centering
\includegraphics[width=0.92\textwidth]{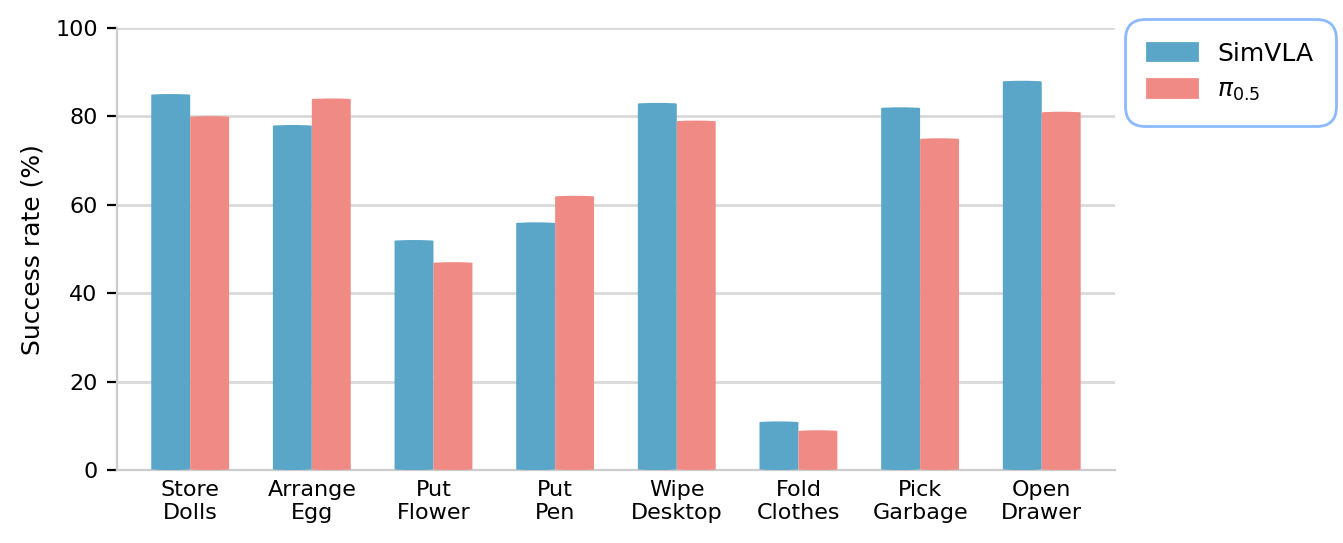}
\caption{\textbf{Real-robot cross-scene zero-shot results on Galaxea R1 Lite.}}
\label{fig:galaxea_zero_shot}
\end{figure}
\section{Conclusion}
\label{sec:conc}

We used \textbf{SimVLA}---a minimal VLM encoder plus lightweight continuous-action head---as a controlled testbed for attributing VLA performance gains. Across LIBERO, CALVIN, WidowX, and Google Robot, the same hierarchy appears: training dynamics have the largest observed effect ($\Delta$$\approx$\crhl{54--94}\%), task configuration ranks second ($\Delta$$\approx$11--43\%), and architecture has the smallest measured effect ($\Delta$$\approx$0.6--15.6\%). SimVLA also achieves leading or competitive performance across the studied benchmarks, showing that careful benchmark-specific training from a pretrained VLM can rival larger VLA-pretrained systems. We therefore propose SimVLA as a matched testbed: future modules should be added while holding training dynamics and task configuration fixed, and gains should be reported as deltas against this baseline. Whether the hierarchy transfers to substantially different VLA families remains open.

\subsection*{Limitations}

SimVLA is intended as a controlled attribution testbed rather than a universal VLA recipe. Its conclusions are therefore strongest for manipulation-oriented VLM-encoder plus lightweight-head policies, where matched training dynamics and task settings can be cleanly compared; Extending the same analysis to broader embodiments, task families, and deployment metrics is an important direction for future work.

\subsection*{Broader Impact}

By clarifying which factors drive VLA performance, SimVLA may make robotic policy research more reproducible and accessible by reducing unnecessary architectural and pretraining costs. At the same time, stronger manipulation policies could accelerate deployment in settings where safety validation, human oversight, and labor impacts require careful assessment. We view SimVLA as an evaluation and ablation tool, not a recommendation for unsupervised deployment in safety-critical environments.

\section*{Acknowledgements and Disclosures}
This work received support from the “111 Center" (No. B26023) and NSFC Grant 62606225.

\bibliographystyle{plainnat}
\bibliography{paper_references}

@article{liu2023libero,
  title={Libero: Benchmarking knowledge transfer for lifelong robot learning},
  author={Liu, Bo and Zhu, Yifeng and Gao, Chongkai and Feng, Yihao and Liu, Qiang and Zhu, Yuke and Stone, Peter},
  journal={Advances in Neural Information Processing Systems},
  volume={36},
  pages={44776--44791},
  year={2023}
}

@inproceedings{xiao2024florence,
  title={Florence-2: Advancing a unified representation for a variety of vision tasks},
  author={Xiao, Bin and Wu, Haiping and Xu, Weijian and Dai, Xiyang and Hu, Houdong and Lu, Yumao and Zeng, Michael and Liu, Ce and Yuan, Lu},
  booktitle={Proceedings of the IEEE/CVF Conference on Computer Vision and Pattern Recognition},
  pages={4818--4829},
  year={2024}
}

@misc{brohan2022rt,
      title={RT-1: Robotics Transformer for Real-World Control at Scale}, 
      author={Anthony Brohan and Noah Brown and Justice Carbajal and Yevgen Chebotar and Joseph Dabis and Chelsea Finn and Keerthana Gopalakrishnan and Karol Hausman and Alex Herzog and Jasmine Hsu and Julian Ibarz and Brian Ichter and Alex Irpan and Tomas Jackson and Sally Jesmonth and Nikhil J Joshi and Ryan Julian and Dmitry Kalashnikov and Yuheng Kuang and Isabel Leal and Kuang-Huei Lee and Sergey Levine and Yao Lu and Utsav Malla and Deeksha Manjunath and Igor Mordatch and Ofir Nachum and Carolina Parada and Jodilyn Peralta and Emily Perez and Karl Pertsch and Jornell Quiambao and Kanishka Rao and Michael Ryoo and Grecia Salazar and Pannag Sanketi and Kevin Sayed and Jaspiar Singh and Sumedh Sontakke and Austin Stone and Clayton Tan and Huong Tran and Vincent Vanhoucke and Steve Vega and Quan Vuong and Fei Xia and Ted Xiao and Peng Xu and Sichun Xu and Tianhe Yu and Brianna Zitkovich},
      year={2023},
      eprint={2212.06817},
      archivePrefix={arXiv},
      primaryClass={cs.RO},
      url={https://arxiv.org/abs/2212.06817}, 
}

@article{zhou2025libero,
  title={LIBERO-PRO: Towards Robust and Fair Evaluation of Vision-Language-Action Models Beyond Memorization},
  author={Zhou, Xueyang and Xu, Yangming and Tie, Guiyao and Chen, Yongchao and Zhang, Guowen and Chu, Duanfeng and Zhou, Pan and Sun, Lichao},
  journal={arXiv preprint arXiv:2510.03827},
  year={2025}
}

@misc{li2024evaluating,
      title={Evaluating Real-World Robot Manipulation Policies in Simulation}, 
      author={Xuanlin Li and Kyle Hsu and Jiayuan Gu and Karl Pertsch and Oier Mees and Homer Rich Walke and Chuyuan Fu and Ishikaa Lunawat and Isabel Sieh and Sean Kirmani and Sergey Levine and Jiajun Wu and Chelsea Finn and Hao Su and Quan Vuong and Ted Xiao},
      year={2024},
      eprint={2405.05941},
      archivePrefix={arXiv},
      primaryClass={cs.RO},
      url={https://arxiv.org/abs/2405.05941}, 
}

@misc{marafioti2025smolvlm,
      title={SmolVLM: Redefining small and efficient multimodal models}, 
      author={Andrés Marafioti and Orr Zohar and Miquel Farré and Merve Noyan and Elie Bakouch and Pedro Cuenca and Cyril Zakka and Loubna Ben Allal and Anton Lozhkov and Nouamane Tazi and Vaibhav Srivastav and Joshua Lochner and Hugo Larcher and Mathieu Morlon and Lewis Tunstall and Leandro von Werra and Thomas Wolf},
      year={2025},
      eprint={2504.05299},
      archivePrefix={arXiv},
      primaryClass={cs.AI},
      url={https://arxiv.org/abs/2504.05299}, 
}

@article{lipman2022flow,
  title={Flow matching for generative modeling},
  author={Lipman, Yaron and Chen, Ricky TQ and Ben-Hamu, Heli and Nickel, Maximilian and Le, Matt},
  journal={arXiv preprint arXiv:2210.02747},
  year={2022}
}

@article{mees2022calvin,
  title={Calvin: A benchmark for language-conditioned policy learning for long-horizon robot manipulation tasks},
  author={Mees, Oier and Hermann, Lukas and Rosete-Beas, Erick and Burgard, Wolfram},
  journal={IEEE Robotics and Automation Letters},
  volume={7},
  number={3},
  pages={7327--7334},
  year={2022},
  publisher={IEEE}
}

@article{kim2024openvla,
  title={OpenVLA: An Open-Source {Vision-Language-Action} Model},
  author={Kim, Moo Jin and Pertsch, Karl and Karamcheti, Siddharth and Xiao, Ted and Balakrishna, Ashwin and Nair, Suraj and Rafailov, Rafael and Foster, Ethan and Lam, Grace and Sanketi, Pannag and Vuong, Quan and Kollar, Thomas and Burchfiel, Benjamin and Tedrake, Russ and Sadigh, Dorsa and Levine, Sergey and Liang, Percy and Finn, Chelsea},
  journal={arXiv preprint arXiv:2406.09246},
  year={2024}
}

@inproceedings{vaswani2017attention,
  title={Attention Is All You Need},
  author={Vaswani, Ashish and Shazeer, Noam and Parmar, Niki and Uszkoreit, Jakob and Jones, Llion and Gomez, Aidan N. and Kaiser, Lukasz and Polosukhin, Illia},
  booktitle={Advances in Neural Information Processing Systems (NeurIPS)},
  year={2017}
}

@misc{liu2025towards,
      title={What Matters in Building Vision-Language-Action Models for Generalist Robots}, 
      author={Xinghang Li and Peiyan Li and Long Qian and Minghuan Liu and Dong Wang and Jirong Liu and Bingyi Kang and Xiao Ma and Xinlong Wang and Di Guo and Tao Kong and Hanbo Zhang and Huaping Liu},
      year={2024},
      eprint={2412.14058},
      archivePrefix={arXiv},
      primaryClass={cs.RO},
      url={https://arxiv.org/abs/2412.14058}, 
}

@misc{walke2023bridgedata,
      title={BridgeData V2: A Dataset for Robot Learning at Scale}, 
      author={Homer Walke and Kevin Black and Abraham Lee and Moo Jin Kim and Max Du and Chongyi Zheng and Tony Zhao and Philippe Hansen-Estruch and Quan Vuong and Andre He and Vivek Myers and Kuan Fang and Chelsea Finn and Sergey Levine},
      year={2024},
      eprint={2308.12952},
      archivePrefix={arXiv},
      primaryClass={cs.RO},
      url={https://arxiv.org/abs/2308.12952}, 
}

@misc{o2024open,
      title={Open X-Embodiment: Robotic Learning Datasets and RT-X Models}, 
      author={Embodiment Collaboration and Abby O'Neill and Abdul Rehman and Abhinav Gupta and Abhiram Maddukuri and Abhishek Gupta and Abhishek Padalkar and Abraham Lee and Acorn Pooley and Agrim Gupta and Ajay Mandlekar and Ajinkya Jain and Albert Tung and Alex Bewley and Alex Herzog and Alex Irpan and Alexander Khazatsky and Anant Rai and Anchit Gupta and Andrew Wang and Andrey Kolobov and Anikait Singh and Animesh Garg and Aniruddha Kembhavi and Annie Xie and Anthony Brohan and Antonin Raffin and Archit Sharma and Arefeh Yavary and Arhan Jain and Ashwin Balakrishna and Ayzaan Wahid and Ben Burgess-Limerick and Beomjoon Kim and Bernhard Schölkopf and Blake Wulfe and Brian Ichter and Cewu Lu and Charles Xu and Charlotte Le and Chelsea Finn and Chen Wang and Chenfeng Xu and Cheng Chi and Chenguang Huang and Christine Chan and Christopher Agia and Chuer Pan and Chuyuan Fu and Coline Devin and Danfei Xu and Daniel Morton and Danny Driess and Daphne Chen and Deepak Pathak and Dhruv Shah and Dieter Büchler and Dinesh Jayaraman and Dmitry Kalashnikov and Dorsa Sadigh and Edward Johns and Ethan Foster and Fangchen Liu and Federico Ceola and Fei Xia and Feiyu Zhao and Felipe Vieira Frujeri and Freek Stulp and Gaoyue Zhou and Gaurav S. Sukhatme and Gautam Salhotra and Ge Yan and Gilbert Feng and Giulio Schiavi and Glen Berseth and Gregory Kahn and Guangwen Yang and Guanzhi Wang and Hao Su and Hao-Shu Fang and Haochen Shi and Henghui Bao and Heni Ben Amor and Henrik I Christensen and Hiroki Furuta and Homanga Bharadhwaj and Homer Walke and Hongjie Fang and Huy Ha and Igor Mordatch and Ilija Radosavovic and Isabel Leal and Jacky Liang and Jad Abou-Chakra and Jaehyung Kim and Jaimyn Drake and Jan Peters and Jan Schneider and Jasmine Hsu and Jay Vakil and Jeannette Bohg and Jeffrey Bingham and Jeffrey Wu and Jensen Gao and Jiaheng Hu and Jiajun Wu and Jialin Wu and Jiankai Sun and Jianlan Luo and Jiayuan Gu and Jie Tan and Jihoon Oh and Jimmy Wu and Jingpei Lu and Jingyun Yang and Jitendra Malik and João Silvério and Joey Hejna and Jonathan Booher and Jonathan Tompson and Jonathan Yang and Jordi Salvador and Joseph J. Lim and Junhyek Han and Kaiyuan Wang and Kanishka Rao and Karl Pertsch and Karol Hausman and Keegan Go and Keerthana Gopalakrishnan and Ken Goldberg and Kendra Byrne and Kenneth Oslund and Kento Kawaharazuka and Kevin Black and Kevin Lin and Kevin Zhang and Kiana Ehsani and Kiran Lekkala and Kirsty Ellis and Krishan Rana and Krishnan Srinivasan and Kuan Fang and Kunal Pratap Singh and Kuo-Hao Zeng and Kyle Hatch and Kyle Hsu and Laurent Itti and Lawrence Yunliang Chen and Lerrel Pinto and Li Fei-Fei and Liam Tan and Linxi "Jim" Fan and Lionel Ott and Lisa Lee and Luca Weihs and Magnum Chen and Marion Lepert and Marius Memmel and Masayoshi Tomizuka and Masha Itkina and Mateo Guaman Castro and Max Spero and Maximilian Du and Michael Ahn and Michael C. Yip and Mingtong Zhang and Mingyu Ding and Minho Heo and Mohan Kumar Srirama and Mohit Sharma and Moo Jin Kim and Muhammad Zubair Irshad and Naoaki Kanazawa and Nicklas Hansen and Nicolas Heess and Nikhil J Joshi and Niko Suenderhauf and Ning Liu and Norman Di Palo and Nur Muhammad Mahi Shafiullah and Oier Mees and Oliver Kroemer and Osbert Bastani and Pannag R Sanketi and Patrick "Tree" Miller and Patrick Yin and Paul Wohlhart and Peng Xu and Peter David Fagan and Peter Mitrano and Pierre Sermanet and Pieter Abbeel and Priya Sundaresan and Qiuyu Chen and Quan Vuong and Rafael Rafailov and Ran Tian and Ria Doshi and Roberto Martín-Martín and Rohan Baijal and Rosario Scalise and Rose Hendrix and Roy Lin and Runjia Qian and Ruohan Zhang and Russell Mendonca and Rutav Shah and Ryan Hoque and Ryan Julian and Samuel Bustamante and Sean Kirmani and Sergey Levine and Shan Lin and Sherry Moore and Shikhar Bahl and Shivin Dass and Shubham Sonawani and Shubham Tulsiani and Shuran Song and Sichun Xu and Siddhant Haldar and Siddharth Karamcheti and Simeon Adebola and Simon Guist and Soroush Nasiriany and Stefan Schaal and Stefan Welker and Stephen Tian and Subramanian Ramamoorthy and Sudeep Dasari and Suneel Belkhale and Sungjae Park and Suraj Nair and Suvir Mirchandani and Takayuki Osa and Tanmay Gupta and Tatsuya Harada and Tatsuya Matsushima and Ted Xiao and Thomas Kollar and Tianhe Yu and Tianli Ding and Todor Davchev and Tony Z. Zhao and Travis Armstrong and Trevor Darrell and Trinity Chung and Vidhi Jain and Vikash Kumar and Vincent Vanhoucke and Vitor Guizilini and Wei Zhan and Wenxuan Zhou and Wolfram Burgard and Xi Chen and Xiangyu Chen and Xiaolong Wang and Xinghao Zhu and Xinyang Geng and Xiyuan Liu and Xu Liangwei and Xuanlin Li and Yansong Pang and Yao Lu and Yecheng Jason Ma and Yejin Kim and Yevgen Chebotar and Yifan Zhou and Yifeng Zhu and Yilin Wu and Ying Xu and Yixuan Wang and Yonatan Bisk and Yongqiang Dou and Yoonyoung Cho and Youngwoon Lee and Yuchen Cui and Yue Cao and Yueh-Hua Wu and Yujin Tang and Yuke Zhu and Yunchu Zhang and Yunfan Jiang and Yunshuang Li and Yunzhu Li and Yusuke Iwasawa and Yutaka Matsuo and Zehan Ma and Zhuo Xu and Zichen Jeff Cui and Zichen Zhang and Zipeng Fu and Zipeng Lin},
      year={2025},
      eprint={2310.08864},
      archivePrefix={arXiv},
      primaryClass={cs.RO},
      url={https://arxiv.org/abs/2310.08864}, 
}

@article{jiang2025galaxea,
  title={Galaxea open-world dataset and g0 dual-system vla model},
  author={Jiang, Tao and Yuan, Tianyuan and Liu, Yicheng and Lu, Chenhao and Cui, Jianning and Liu, Xiao and Cheng, Shuiqi and Gao, Jiyang and Xu, Huazhe and Zhao, Hang},
  journal={arXiv preprint arXiv:2509.00576},
  year={2025}
}

@misc{bjorck2025gr00t,
      title={GR00T N1: An Open Foundation Model for Generalist Humanoid Robots}, 
      author={NVIDIA and : and Johan Bjorck and Fernando Castañeda and Nikita Cherniadev and Xingye Da and Runyu Ding and Linxi "Jim" Fan and Yu Fang and Dieter Fox and Fengyuan Hu and Spencer Huang and Joel Jang and Zhenyu Jiang and Jan Kautz and Kaushil Kundalia and Lawrence Lao and Zhiqi Li and Zongyu Lin and Kevin Lin and Guilin Liu and Edith Llontop and Loic Magne and Ajay Mandlekar and Avnish Narayan and Soroush Nasiriany and Scott Reed and You Liang Tan and Guanzhi Wang and Zu Wang and Jing Wang and Qi Wang and Jiannan Xiang and Yuqi Xie and Yinzhen Xu and Zhenjia Xu and Seonghyeon Ye and Zhiding Yu and Ao Zhang and Hao Zhang and Yizhou Zhao and Ruijie Zheng and Yuke Zhu},
      year={2025},
      eprint={2503.14734},
      archivePrefix={arXiv},
      primaryClass={cs.RO},
      url={https://arxiv.org/abs/2503.14734}, 
}

@misc{DiffusionPolicy,
      title={Diffusion Policy: Visuomotor Policy Learning via Action Diffusion}, 
      author={Cheng Chi and Zhenjia Xu and Siyuan Feng and Eric Cousineau and Yilun Du and Benjamin Burchfiel and Russ Tedrake and Shuran Song},
      year={2024},
      eprint={2303.04137},
      archivePrefix={arXiv},
      primaryClass={cs.RO},
      url={https://arxiv.org/abs/2303.04137}, 
}

@misc{VLA_anatomy_survey,
      title={An Anatomy of Vision-Language-Action Models: From Modules to Milestones and Challenges}, 
      author={Chao Xu and Suyu Zhang and Yang Liu and Baigui Sun and Weihong Chen and Bo Xu and Qi Liu and Juncheng Wang and Shujun Wang and Shan Luo and Jan Peters and Athanasios V. Vasilakos and Stefanos Zafeiriou and Jiankang Deng},
      year={2025},
      eprint={2512.11362},
      archivePrefix={arXiv},
      primaryClass={cs.RO},
      url={https://arxiv.org/abs/2512.11362}, 
}

@misc{VLA_model_survey,
      title={Vision-Language-Action Models: Concepts, Progress, Applications and Challenges},
      author={Ranjan Sapkota and Yang Cao and Konstantinos I. Roumeliotis and Manoj Karkee},
      year={2025},
      eprint={2505.04769},
      archivePrefix={arXiv},
      primaryClass={cs.CV},
      url={https://arxiv.org/abs/2505.04769}, 
}

@misc{GAI_survey,
      title={Generative Artificial Intelligence in Robotic Manipulation: A Survey},
      author={Kun Zhang and Peng Yun and Jun Cen and Junhao Cai and Didi Zhu and Hangjie Yuan and Chao Zhao and Tao Feng and Michael Yu Wang and Qifeng Chen and Jia Pan and Wei Zhang and Bo Yang and Hua Chen},
      year={2025},
      eprint={2503.03464},
      archivePrefix={arXiv},
      primaryClass={cs.RO},
      url={https://arxiv.org/abs/2503.03464}, 
}

@article{FlowVLA,
  title={FlowVLA: Visual Chain of Thought-based Motion Reasoning for Vision-Language-Action Models},
  author={Zhide Zhong and Haodong Yan and Junfeng Li and Xiangchen Liu and Xin Gong and Tianran Zhang and Wenxuan Song and Jiayi Chen and Xinhu Zheng and Hesheng Wang and Haoang Li},
  journal={arXiv preprint arXiv:2508.18269},
  year={2025}
}

@article{UnifiedVLA,
  title={Unified Vision-Language-Action Model},
  author={Yuqi Wang and Xinghang Li and Wenxuan Wang and Junbo Zhang and Yingyan Li and Yuntao Chen and Xinlong Wang and Zhaoxiang Zhang},
  journal={arXiv preprint arXiv:2506.19850},
  year={2025}
}

@article{OpenVLA_OFT,
  title={Fine-Tuning Vision-Language-Action Models: Optimizing Speed and Success},
  author={Moo Jin Kim and Chelsea Finn and Percy Liang},
  journal={arXiv preprint arXiv:2502.19645},
  year={2025}
}

@article{DD_VLA,
  title={Discrete Diffusion VLA: Bringing Discrete Diffusion to Action Decoding in Vision-Language-Action Policies},
  author={Zhixuan Liang and Yizhuo Li and Tianshuo Yang and Chengyue Wu and Sitong Mao and Tian Nian and Liuao Pei and Shunbo Zhou and Xiaokang Yang and Jiangmiao Pang and Yao Mu and Ping Luo},
  journal={arXiv preprint arXiv:2508.20072},
  year={2025}
}

@article{DepthVLA,
  title={DepthVLA: Enhancing Vision-Language-Action Models with Depth-Aware Spatial Reasoning},
  author={Tianyuan Yuan and Yicheng Liu and Chenhao Lu and Zhuoguang Chen and Tao Jiang and Hang Zhao},
  journal={arXiv preprint arXiv:2510.13375},
  year={2025}
}

@article{PixelVLA,
  title={PixelVLA: Advancing Pixel-level Understanding in Vision-Language-Action Model},
  author={Wenqi Liang and Gan Sun and Yao He and Jiahua Dong and Suyan Dai and Ivan Laptev and Salman Khan and Yang Cong},
  journal={arXiv preprint arXiv:2511.01571},
  year={2025}
}

@article{IntentionVLA,
  title={IntentionVLA: Generalizable and Efficient Embodied Intention Reasoning for Human-Robot Interaction},
  author={Yandu Chen and Kefan Gu and Yuqing Wen and Yucheng Zhao and Tiancai Wang and Liqiang Nie},
  journal={arXiv preprint arXiv:2510.07778},
  year={2025}
}

@article{FPC_VLA,
  title={FPC-VLA: A Vision-Language-Action Framework with a Supervisor for Failure Prediction and Correction},
  author={Yifan Yang and Zhixiang Duan and Tianshi Xie and Fuyu Cao and Pinxi Shen and Peili Song and Piaopiao Jin and Guokang Sun and Shaoqing Xu and Yangwei You and Jingtai Liu},
  journal={arXiv preprint arXiv:2509.04018},
  year={2025}
}

@article{HAMLET,
  title={HAMLET: Switch your Vision-Language-Action Model into a History-Aware Policy},
  author={Myungkyu Koo and Daewon Choi and Taeyoung Kim and Kyungmin Lee and Changyeon Kim and Younggyo Seo and Jinwoo Shin},
  journal={arXiv preprint arXiv:2510.00695},
  year={2025}
}

@article{Dream_VLA,
  title={DreamVLA: A Vision-Language-Action Model Dreamed with Comprehensive World Knowledge},
  author={Wenyao Zhang and Hongsi Liu and Zekun Qi and Yunnan Wang and Xinqiang Yu and Jiazhao Zhang and Runpei Dong and Jiawei He and Fan Lu and He Wang and Zhizheng Zhang and Li Yi and Wenjun Zeng and Xin Jin},
  journal={arXiv preprint arXiv:2507.04447},
  year={2025}
}

@article{NORA,
  title={NORA: A Small Open-Sourced Generalist Vision Language Action Model for Embodied Tasks},
  author={Chia-Yu Hung and Qi Sun and Pengfei Hong and Amir Zadeh and Chuan Li and U-Xuan Tan and Navonil Majumder and Soujanya Poria},
  journal={arXiv preprint arXiv:2504.19854},
  year={2025}
}

@article{FLOWER,
  title={FLOWER: Democratizing Generalist Robot Policies with Efficient Vision-Language-Action Flow Policies},
  author={Moritz Reuss and Hongyi Zhou and Marcel Rühle and Ömer Erdinç Yağmurlu and Fabian Otto and Rudolf Lioutikov},
  journal={arXiv preprint arXiv:2509.04996},
  year={2025}
}

@article{VLA_Adapter,
  title={VLA-Adapter: An Effective Paradigm for Tiny-Scale Vision-Language-Action Model},
  author={Yihao Wang and Pengxiang Ding and Lingxiao Li and Can Cui and Zirui Ge and Xinyang Tong and Wenxuan Song and Han Zhao and Wei Zhao and Pengxu Hou and Siteng Huang and Yifan Tang and Wenhui Wang and Ru Zhang and Jianyi Liu and Donglin Wang},
  journal={arXiv preprint arXiv:2509.09372},
  year={2025}
}

@article{GeoVLA,
  title={GeoVLA: Empowering 3D Representations in Vision-Language-Action Models},
  author={Lin Sun and Bin Xie and Yingfei Liu and Hao Shi and Tiancai Wang and Jiale Cao},
  journal={arXiv preprint arXiv:2508.09071},
  year={2025}
}

@article{FALCON,
  title={From Spatial to Actions: Grounding Vision-Language-Action Model in Spatial Foundation Priors},
  author={Zhengshen Zhang and Hao Li and Yalun Dai and Zhengbang Zhu and Lei Zhou and Chenchen Liu and Dong Wang and Francis E. H. Tay and Sijin Chen and Ziwei Liu and Yuxiao Liu and Xinghang Li and Pan Zhou},
  journal={arXiv preprint arXiv:2510.17439},
  year={2025}
}

@article{x_VLA,
  title={X-VLA: Soft-Prompted Transformer as Scalable Cross-Embodiment Vision-Language-Action Model},
  author={Jinliang Zheng and Jianxiong Li and Zhihao Wang and Dongxiu Liu and Xirui Kang and Yuchun Feng and Yinan Zheng and Jiayin Zou and Yilun Chen and Jia Zeng and Ya-Qin Zhang and Jiangmiao Pang and Jingjing Liu and Tai Wang and Xianyuan Zhan},
  journal={arXiv preprint arXiv:2510.10274},
  year={2025}
}

@misc{PD_VLA,
      title={PD-VLA: Accelerating Vision-Language-Action Model Integrated with Action Chunking via Parallel Decoding}, 
      author={Wenxuan Song and Jiayi Chen and Pengxiang Ding and Han Zhao and Wei Zhao and Zhide Zhong and Zongyuan Ge and Zhijun Li and Donglin Wang and Jun Ma and Lujia Wang and Haoang Li},
      year={2025},
      eprint={2503.02310},
      archivePrefix={arXiv},
      primaryClass={cs.RO},
      url={https://arxiv.org/abs/2503.02310}, 
}

@article{CoT_VLA,
  title={CoT-VLA: Visual Chain-of-Thought Reasoning for Vision-Language-Action Models},
  author={Qingqing Zhao and Yao Lu and Moo Jin Kim and Zipeng Fu and Zhuoyang Zhang and Yecheng Wu and Zhaoshuo Li and Qianli Ma and Song Han and Chelsea Finn and Ankur Handa and Ming-Yu Liu and Donglai Xiang and Gordon Wetzstein and Tsung-Yi Lin},
  journal={CVPR 2025},
  year={2025},
  note={arXiv:2503.22020}
}

@article{WorldVLA,
  title={WorldVLA: Towards Autoregressive Action World Model},
  author={Jun Cen and Chaohui Yu and Hangjie Yuan and Yuming Jiang and Siteng Huang and Jiayan Guo and Xin Li and Yibing Song and Hao Luo and Fan Wang and Deli Zhao and Hao Chen},
  journal={arXiv preprint arXiv:2506.21539},
  year={2025}
}

@Article{TraceVLA,
 author = {Ruijie Zheng and Yongyuan Liang and Shuaiyi Huang and Jianfeng Gao and Hal Daum'e and Andrey Kolobov and Furong Huang and Jianwei Yang},
 booktitle = {International Conference on Learning Representations},
 journal = {ArXiv},
 title = {TraceVLA: Visual Trace Prompting Enhances Spatial-Temporal Awareness for Generalist Robotic Policies},
 volume = {abs/2412.10345},
 year = {2024}
}

@article{p4D_VLA,
  title={4D-VLA: Spatiotemporal Vision-Language-Action Pretraining with Cross-Scene Calibration},
  author={Jiahui Zhang and Yurui Chen and Yueming Xu and Ze Huang and Yanpeng Zhou and Yu-Jie Yuan and Xinyue Cai and Guowei Huang and Xingyue Quan and Hang Xu and Li Zhang},
  journal={arXiv preprint arXiv:2506.22242},
  year={2025}
}

@article{GraspVLA,
  title={GraspVLA: a Grasping Foundation Model Pre-trained on Billion-scale Synthetic Action Data},
  author={Shengliang Deng and Mi Yan and Songlin Wei and Haixin Ma and Yuxin Yang and Jiayi Chen and Zhiqi Zhang and Taoyu Yang and Xuheng Zhang and Wenhao Zhang and Heming Cui and Zhizheng Zhang and He Wang},
  journal={arXiv preprint arXiv:2505.03233},
  year={2025}
}

@article{VLA_OS,
  title={VLA-OS: Structuring and Dissecting Planning Representations and Paradigms in Vision-Language-Action Models},
  author={Chongkai Gao and Zixuan Liu and Zhenghao Chi and Junshan Huang and Xin Fei and Yiwen Hou and Yuxuan Zhang and Yudi Lin and Zhirui Fang and Zeyu Jiang and Lin Shao},
  journal={arXiv preprint arXiv:2506.17561},
  year={2025}
}

@misc{Octo,
      title={Octo: An Open-Source Generalist Robot Policy}, 
      author={Octo Model Team and Dibya Ghosh and Homer Walke and Karl Pertsch and Kevin Black and Oier Mees and Sudeep Dasari and Joey Hejna and Tobias Kreiman and Charles Xu and Jianlan Luo and You Liang Tan and Lawrence Yunliang Chen and Pannag Sanketi and Quan Vuong and Ted Xiao and Dorsa Sadigh and Chelsea Finn and Sergey Levine},
      year={2024},
      eprint={2405.12213},
      archivePrefix={arXiv},
      primaryClass={cs.RO},
      url={https://arxiv.org/abs/2405.12213}, 
}

@article{SmolVLA,
  title={SmolVLA: A Vision-Language-Action Model for Affordable and Efficient Robotics},
  author={Mustafa Shukor and Dana Aubakirova and Francesco Capuano and Pepijn Kooijmans and Steven Palma and Adil Zouitine and Michel Aractingi and Caroline Pascal and Martino Russi and Andres Marafioti and Simon Alibert and Matthieu Cord and Thomas Wolf and Remi Cadene},
  journal={arXiv preprint arXiv:2506.01844},
  year={2025}
}

@article{VQ_VLA,
  title={VQ-VLA: Improving Vision-Language-Action Models via Scaling Vector-Quantized Action Tokenizers},
  author={Yating Wang and Haoyi Zhu and Mingyu Liu and Jiange Yang and Hao-Shu Fang and Tong He},
  journal={arXiv preprint arXiv:2507.01016},
  year={2025}
}

@article{pi0,
  title={{\(\pi_0\): A Vision-Language-Action Flow Model for General Robot Control}},
  author={Kevin Black and Noah Brown and Danny Driess and Adnan Esmail and Michael Equi and Chelsea Finn and Niccolo Fusai and Lachy Groom and Karol Hausman and Brian Ichter and Szymon Jakubczak and Tim Jones and Liyiming Ke and Sergey Levine and Adrian Li-Bell and Mohith Mothukuri and Suraj Nair and Karl Pertsch and Lucy Xiaoyang Shi and James Tanner and Quan Vuong and Anna Walling and Haohuan Wang and Ury Zhilinsky},
  journal={arXiv preprint arXiv:2410.24164},
  year={2024}
}

@article{FAST,
  title={FAST: Efficient Action Tokenization for Vision-Language-Action Models},
  author={Karl Pertsch and Kyle Stachowicz and Brian Ichter and Danny Driess and Suraj Nair and Quan Vuong and Oier Mees and Chelsea Finn and Sergey Levine},
  journal={arXiv preprint arXiv:2501.09747},
  year={2025}
}

@article{pi05,
  title={{\(\pi_{0.5}\): a Vision-Language-Action Model with Open-World Generalization}},
  author={Kevin Black and Noah Brown and James Darpinian and Karan Dhabalia and Danny Driess and Adnan Esmail and Michael Equi and Chelsea Finn and Niccolo Fusai and Manuel Y. Galliker and Dibya Ghosh and Lachy Groom and Karol Hausman and Brian Ichter and Szymon Jakubczak and Tim Jones and Liyiming Ke and Devin LeBlanc and Sergey Levine and Adrian Li-Bell and Mohith Mothukuri and Suraj Nair and Karl Pertsch and Allen Z. Ren and Lucy Xiaoyang Shi and Laura Smith and Jost Tobias Springenberg and Kyle Stachowicz and James Tanner and Quan Vuong and Homer Walke and Anna Walling and Haohuan Wang and Lili Yu and Ury Zhilinsky},
  journal={arXiv preprint arXiv:2504.16054},
  year={2025}
}

@article{ReconVLA,
  title={ReconVLA: Reconstructive Vision-Language-Action Model as Effective Robot Perceiver},
  author={Wenxuan Song and Ziyang Zhou and Han Zhao and Jiayi Chen and Pengxiang Ding and Haodong Yan and Yuxin Huang and Feilong Tang and Donglin Wang and Haoang Li},
  journal={arXiv preprint arXiv:2508.10333},
  year={2025}
}

@article{UniAct,
  title={Universal Actions for Enhanced Embodied Foundation Models},
  author={Jinliang Zheng and Jianxiong Li and Dongxiu Liu and Yinan Zheng and Zhihao Wang and Zhonghong Ou and Yu Liu and Jingjing Liu and Ya-Qin Zhang and Xianyuan Zhan},
  journal={arXiv preprint arXiv:2501.10105},
  year={2025}
}

@article{ThinkAct,
  title={ThinkAct: Vision-Language-Action Reasoning via Reinforced Visual Latent Planning},
  author={Chi-Pin Huang and Yueh-Hua Wu and Min-Hung Chen and Yu-Chiang Frank Wang and Fu-En Yang},
  journal={arXiv preprint arXiv:2507.16815},
  year={2025}
}

@article{MolmoAct,
  title={MolmoAct: Action Reasoning Models that can Reason in Space},
  author={Jason Lee and Jiafei Duan and Haoquan Fang and Yuquan Deng and Shuo Liu and Boyang Li and Bohan Fang and Jieyu Zhang and Yi Ru Wang and Sangho Lee and Winson Han and Wilbert Pumacay and Angelica Wu and Rose Hendrix and Karen Farley and Eli VanderBilt and Ali Farhadi and Dieter Fox and Ranjay Krishna},
  journal={arXiv preprint arXiv:2508.07917},
  year={2025}
}

@article{MemoryVLA,
  title={MemoryVLA: Perceptual-Cognitive Memory in Vision-Language-Action Models for Robotic Manipulation},
  author={Hao Shi and Bin Xie and Yingfei Liu and Lin Sun and Fengrong Liu and Tiancai Wang and Erjin Zhou and Haoqiang Fan and Xiangyu Zhang and Gao Huang},
  journal={arXiv preprint arXiv:2508.19236},
  year={2025}
}

@article{UniVLA,
  title={UniVLA: Learning to Act Anywhere with Task-centric Latent Actions},
  author={Qingwen Bu and Yanting Yang and Jisong Cai and Shenyuan Gao and Guanghui Ren and Maoqing Yao and Ping Luo and Hongyang Li},
  journal={arXiv preprint arXiv:2505.06111},
  year={2025}
}

@misc{SpatialVLA,
 title={SpatialVLA: Exploring Spatial Representations for Visual-Language-Action Model},
 author={Delin Qu and Haoming Song and Qizhi Chen and Yuanqi Yao and Xinyi Ye and Yan Ding and Zhigang Wang and JiaYuan Gu and Bin Zhao and Dong Wang and Xuelong Li},
 year={2025},
 eprint={2501.15830},
 archivePrefix={arXiv},
 primaryClass={cs.RO},
 url={https://arxiv.org/abs/2501.15830},
}

@misc{BEHAVIOR1K,
      title={BEHAVIOR-1K: A Human-Centered, Embodied AI Benchmark with 1,000 Everyday Activities and Realistic Simulation}, 
      author={Chengshu Li and Ruohan Zhang and Josiah Wong and Cem Gokmen and Sanjana Srivastava and Roberto Martín-Martín and Chen Wang and Gabrael Levine and Wensi Ai and Benjamin Martinez and Hang Yin and Michael Lingelbach and Minjune Hwang and Ayano Hiranaka and Sujay Garlanka and Arman Aydin and Sharon Lee and Jiankai Sun and Mona Anvari and Manasi Sharma and Dhruva Bansal and Samuel Hunter and Kyu-Young Kim and Alan Lou and Caleb R Matthews and Ivan Villa-Renteria and Jerry Huayang Tang and Claire Tang and Fei Xia and Yunzhu Li and Silvio Savarese and Hyowon Gweon and C. Karen Liu and Jiajun Wu and Li Fei-Fei},
      year={2024},
      eprint={2403.09227},
      archivePrefix={arXiv},
      primaryClass={cs.RO},
      url={https://arxiv.org/abs/2403.09227}, 
}

@misc{larchenko2025behavior,
  title={Task adaptation of Vision-Language-Action model: 1st Place Solution for the 2025 BEHAVIOR Challenge},
  author={Larchenko, Ilia and Zarin, Gleb and Karnatak, Akash},
  year={2025},
  eprint={2512.06951},
  archivePrefix={arXiv},
  primaryClass={cs.RO},
  url={https://arxiv.org/abs/2512.06951}
}

@misc{wu2026vlanext,
      title={VLANeXt: Recipes for Building Strong VLA Models}, 
      author={Xiao-Ming Wu and Bin Fan and Kang Liao and Jian-Jian Jiang and Runze Yang and Yihang Luo and Zhonghua Wu and Wei-Shi Zheng and Chen Change Loy},
      year={2026},
      eprint={2602.18532},
      archivePrefix={arXiv},
      primaryClass={cs.CV},
      url={https://arxiv.org/abs/2602.18532}, 
}

@misc{hoffmann2022training,
      title={Training Compute-Optimal Large Language Models}, 
      author={Jordan Hoffmann and Sebastian Borgeaud and Arthur Mensch and Elena Buchatskaya and Trevor Cai and Eliza Rutherford and Diego de Las Casas and Lisa Anne Hendricks and Johannes Welbl and Aidan Clark and Tom Hennigan and Eric Noland and Katie Millican and George van den Driessche and Bogdan Damoc and Aurelia Guy and Simon Osindero and Karen Simonyan and Erich Elsen and Jack W. Rae and Oriol Vinyals and Laurent Sifre},
      year={2022},
      eprint={2203.15556},
      archivePrefix={arXiv},
      primaryClass={cs.CL},
      url={https://arxiv.org/abs/2203.15556}, 
}

@misc{touvron2023llama,
      title={LLaMA: Open and Efficient Foundation Language Models}, 
      author={Hugo Touvron and Thibaut Lavril and Gautier Izacard and Xavier Martinet and Marie-Anne Lachaux and Timothée Lacroix and Baptiste Rozière and Naman Goyal and Eric Hambro and Faisal Azhar and Aurelien Rodriguez and Armand Joulin and Edouard Grave and Guillaume Lample},
      year={2023},
      eprint={2302.13971},
      archivePrefix={arXiv},
      primaryClass={cs.CL},
      url={https://arxiv.org/abs/2302.13971}, 
}

@article{HiF-VLA,
  title={HiF-VLA: Hindsight, Insight and Foresight through Motion Representation for Vision-Language-Action Models},
  author={Lin, Minghui and Ding, Pengxiang and Wang, Shu and Zhuang, Zifeng and Liu, Yang and Tong, Xinyang and Song, Wenxuan and Lyu, Shangke and Huang, Siteng and Wang, Donglin},
  journal={arXiv preprint arXiv:2512.09928},
  year={2025}
}

@article{Seer,
  title={Predictive inverse dynamics models are scalable learners for robotic manipulation},
  author={Tian, Yang and Yang, Sizhe and Zeng, Jia and Wang, Ping and Lin, Dahua and Dong, Hao and Pang, Jiangmiao},
  journal={arXiv preprint arXiv:2412.15109},
  year={2024}
}

@misc{qwen3vl,
      title={Qwen3-VL Technical Report},
      author={{Qwen Team}},
      year={2025},
      note={TODO: confirm final citation source before submission},
}

\newpage
\appendix
\section*{Appendix}
\label{ap-a}

\section{Baselines}\label{app:baselines}
\paragraph{Benchmark comparison baselines.}
We compare SimVLA against a spectrum of 34 representative policies in the benchmark tables, ranging from standard baselines to recent complex architectures: RT-1-X / RT-2-X \citep{o2024open}, OpenVLA \citep{kim2024openvla}, Octo-Small / Octo-Base \citep{Octo}, TraceVLA \citep{TraceVLA}, SpatialVLA \citep{SpatialVLA}, UnifiedVLA \citep{UnifiedVLA}, UniVLA \citep{UniVLA}, X-VLA \citep{x_VLA}, VLA-Adapter \citep{VLA_Adapter}, VLA-OS \citep{VLA_OS}, UniAct \citep{UniAct}, NORA \citep{NORA}, MemoryVLA \citep{MemoryVLA}, ThinkAct \citep{ThinkAct}, CoT-VLA \citep{CoT_VLA}, WorldVLA \citep{WorldVLA}, SmolVLA \citep{SmolVLA}, MolmoAct \citep{MolmoAct}, $\pi_0$ \citep{pi0}, $\pi_0$-FAST \citep{FAST}, $\pi_{0.5}$ \citep{pi05}, DD-VLA \citep{DD_VLA}, OpenVLA-OFT \citep{OpenVLA_OFT}, RoboVLM \citep{liu2025towards}, GR00T-N1 \citep{bjorck2025gr00t}, FlowVLA \citep{FlowVLA}, PD-VLA \citep{PD_VLA}, FPC-VLA \citep{FPC_VLA}, 4D-VLA \citep{p4D_VLA}, GraspVLA \citep{GraspVLA}, FLOWER \citep{FLOWER}, VLANeXt \citep{wu2026vlanext}, HiF-VLA \citep{HiF-VLA}, Seer \citep{Seer}.


\paragraph{Related-work categorization.}\label{app:baseline_categorization}
Table~\ref{tab:baseline_categorization} maps the baselines discussed in Section~\ref{sec:related_work} to their corresponding design axis. We organize existing VLA methods along three axes: (1)~\emph{Visual and Temporal Augmentation}, which adds auxiliary visual cues such as flow, traces, or chain-of-thought reasoning; (2)~\emph{Geometric and 3D Priors}, which injects spatial structure via depth, point clouds, or 3D representations; and (3)~\emph{Action Representations and Architectures}, which focuses on the action decoding mechanism itself (e.g., diffusion, flow matching, autoregressive tokenization, or adapter-based heads). The majority of recent baselines fall into Category~3, reflecting the community's focus on action-head design. SimVLA achieves leading performance without the auxiliary modules from Categories~1 and~2, suggesting that a well-tuned training recipe applied to a simple architecture can match or exceed methods that rely on additional visual or spatial priors.

\begin{table}[h!]
\caption{\textbf{Categorization Summary.} The baselines are mapped to their corresponding design axis as defined in Section~\ref{sec:related_work} (Related Work).}
\label{tab:baseline_categorization}
\centering
\small
\begin{tabular}{p{0.30\linewidth}|p{0.63\linewidth}}
\toprule
\textbf{Design Axis} & \textbf{Models} \\
\midrule
\textit{(1) Visual and Temporal Augmentation} &
\begin{minipage}[t]{\linewidth}\raggedright
FlowVLA~\citep{FlowVLA}, CoT-VLA~\citep{CoT_VLA}, \\
TraceVLA~\citep{TraceVLA}, 4D-VLA~\citep{p4D_VLA}, \\
ThinkAct~\citep{ThinkAct}, MemoryVLA~\citep{MemoryVLA}, FPC-VLA~\citep{FPC_VLA}
\end{minipage} \\
\midrule
\textit{(2) Geometric and 3D Priors} &
\begin{minipage}[t]{\linewidth}\raggedright
MolmoAct~\citep{MolmoAct}, SpatialVLA~\citep{SpatialVLA}, \\
GraspVLA~\citep{GraspVLA}, 4D-VLA~\citep{p4D_VLA}
\end{minipage} \\
\midrule
\textit{(3) Action Representations and Architectures} &
\begin{minipage}[t]{\linewidth}\raggedright
OpenVLA-OFT~\citep{OpenVLA_OFT}, DD-VLA~\citep{DD_VLA}, \\
PD-VLA~\citep{PD_VLA}, UniVLA~\citep{UniVLA}, \\
UnifiedVLA~\citep{UnifiedVLA}, UniAct~\citep{UniAct}, $\pi_0$~\citep{pi0}, \\
$\pi_0$-FAST~\citep{FAST}, $\pi_{0.5}$~\citep{pi05}, NORA~\citep{NORA}, \\
SmolVLA~\citep{SmolVLA}, GR00T-N1~\citep{bjorck2025gr00t}, \\
FLOWER~\citep{FLOWER}, X-VLA~\citep{x_VLA}, VLA-Adapter~\citep{VLA_Adapter}, \\
VLA-OS~\citep{VLA_OS}, WorldVLA~\citep{WorldVLA}
\end{minipage} \\
\bottomrule
\end{tabular}
\end{table}

\section{Galaxea Open-World Dataset}\label{app:galaxea_dataset}
We train on the Galaxea Open-World Dataset \citep{jiang2025galaxea}, a large-scale real-world mobile-manipulation dataset with $\sim$500 hours of demonstrations (about 100K trajectories). The dataset spans 150 task categories across 50 real-world scenes and covers more than 1{,}600 objects and 58 skills. Importantly for our setting, data are collected under a single consistent embodiment, so that perception streams, action/state signals, and language annotations are naturally aligned for end-to-end VLA training.

\paragraph{Platform.}
All demonstrations are recorded on the Galaxea R1 Lite platform, which is also the real robot we use for our cross-scene zero-shot evaluation in Sec.~\ref{sec:evaluation}. R1 Lite is a mobile bimanual robot with a 23-DoF embodiment (two 6-DoF arms, a 3-DoF torso for workspace extension, a 6-DoF vector-drive omnidirectional base up to 1.5\,m/s, and two 1-DoF grippers). The platform is equipped with a head stereo RGB camera for scene-level context and dual Intel RealSense D405 RGB-D wrist cameras for close-range manipulation.

\section{Real-Robot Evaluation Details}\label{app:real_robot_eval_details}
For each real-robot task, we use a simple rubric that measures whether the policy can complete the task within the fixed time budget used in Sec.~\ref{sec:evaluation}. We report binary success/failure over 50 trials.

\paragraph{Store the dolls.}
The robot must pick up \textbf{three} plush dolls placed on a tabletop and put them into a designated storage container. Success is defined as all three dolls being fully inside the container at the end of the episode.

\paragraph{Arrange eggs.}
The robot must pick up \textbf{one} egg and place it into a designated slot of an egg carton. Success is defined as the egg being seated in the target slot without being dropped.

\paragraph{Put the flowers in the vase.}
The robot must grasp a flower and insert it into the opening of a vase. Success is defined as the flower being inside the vase and remaining stably placed at the end of the episode.

\paragraph{Put the pen into the pen holder.}
The robot must pick up \textbf{two} pens and insert them into a pen holder. Success is defined as both pens being inside the holder (stably contained) at the end of the episode.

\paragraph{Wipe the desktop.}
The robot must wipe a visible stain/dirty region on a desktop using a \textbf{cloth rag}. Success is defined as the target region being visibly cleaned (stain disappears or is substantially reduced) within the time budget.

\paragraph{Fold the clothes.}
The robot must fold a shirt starting from a flat or mildly wrinkled state on the table. Success is defined as completing the target fold (folding in sleeves and making a main fold) with the garment remaining on the working surface.

\paragraph{Pick up garbage on the ground.}
The robot must pick up \textbf{two} plastic bottles from the ground and dispose them into a trash bin. Success is defined as both bottles ending inside the bin.

\paragraph{Open the drawer.}
The robot must grasp the drawer handle and pull to open the drawer to a specified extent. Success is defined as the drawer being opened beyond a threshold without damaging the mechanism.

\section{Training Hyperparameters}\label{app:training_hparams}
To facilitate reproducibility, we summarize the key hyperparameters used in our simulation training (LIBERO, CALVIN, WidowX, Google Robot, and BEHAVIOR-1K) and real-robot training (Galaxea-500h). All simulation runs are executed on \textbf{4$\times$H100} GPUs, except BEHAVIOR-1K which uses \textbf{8$\times$H100}, and are initialized from a pretrained VLM backbone without additional large-scale robot-data VLA pretraining. For comprehensive implementation details, please refer to our codebase.

\paragraph{Simulation.}
Table~\ref{tab:hparams_libero} reports the configuration.

\begin{table}[t]
\centering
\caption{\textbf{Training Hyperparameters and Inference Settings for Simulation Datasets.}}
\label{tab:hparams_libero}
\scriptsize
\resizebox{\textwidth}{!}{%
\begin{tabular}{l|lllll}
\toprule
\textbf{Configuration} & \textbf{Libero} & \textbf{WidowX} & \textbf{Google Robot}  & \textbf{Calvin} & \textbf{BEHAVIOR-1K} \\
\midrule
Batch size (per GPU) & 64 & 80 & 80 & 48 & 32\\
Global batch size & 64$\times$4 = 256 & 80$\times$4 = 320 & 80$\times$4 = 320 & 48$\times$4 = 192 & 32$\times$8 = 256\\
Action head & Large (1024,24,16) & Large (1024,24,16) & Large (1024,24,16) & Large (1024,24,16) & Large (1024,24,16)\\
Action chunk horizon \(H\) & 10 & 30 & 30 & 30 & 30\\
Action objective & conditional flow matching & conditional flow matching & conditional flow matching & conditional flow matching & conditional flow matching\\
Noise distribution & standard Gaussian & standard Gaussian & standard Gaussian & standard Gaussian & standard Gaussian\\
Time sampling & $\text{Beta}(1.5,1)$ on $[0.001,1]$ & $\text{Beta}(1.5,1)$ on $[0.001,1]$ & $\text{Beta}(1.5,1)$ on $[0.001,1]$ & $\text{Beta}(1.5,1)$ on $[0.001,1]$ & $\text{Beta}(1.5,1)$ on $[0.001,1]$\\
Training target & velocity \((\epsilon-x)\) & velocity \((\epsilon-x)\) & velocity \((\epsilon-x)\) & velocity \((\epsilon-x)\) & velocity \((\epsilon-x)\)\\
Inference solver & Euler integration & Euler integration & Euler integration & Euler integration & Euler integration\\
Inference steps & 10 & 10 & 10 & 10 & 10\\
Image resize & 128$\times$128 & 224$\times$224 & 224$\times$224 & 224$\times$224 & 224$\times$224\\
Action normalization & on & on & on & on & on\\
Data shuffling & on & on & on & on & on\\
Optimizer & AdamW & AdamW & AdamW & AdamW & AdamW\\
Betas & (0.9, 0.95) & (0.9, 0.95) & (0.9, 0.95) & (0.9, 0.95) & (0.9, 0.95)\\
Weight decay & 0.0 & 0.0 & 0.0 & 0.0 & 0.0\\
Learning rate & 2e-4 & 1e-4 & 1e-4 & 1e-4 & 1e-4\\
VLM LR multiplier & 0.1 & 0.05 & 0.01 & 0.1 & 0.01\\
Warm-up steps & 0 & 2000 & 2000 & 2000 & 2000\\
Scheduler & none & none & none & cosine & none\\
Training steps & 150K & 100K & 150K & 150K & 300K\\
Precision & bfloat16 & bfloat16 & bfloat16 & bfloat16 & bfloat16\\
\bottomrule
\end{tabular}
}
\end{table}

\paragraph{Real-robot training (Galaxea-500h).}
Table~\ref{tab:hparams_galaxea} summarizes the training setup on the Galaxea Open-World Dataset. For a fair comparison, both our method and $\pi_{0.5}$ are trained using the same Galaxea data and the same hyperparameters, with the same compute budget: \textbf{64$\times$H100} GPUs for \textbf{150K} training steps.

\begin{table}[h]
\centering
\vspace{-0.0 in}
\caption{\textbf{Hyperparameters for Galaxea-500h Training.}}
\label{tab:hparams_galaxea}
\scriptsize
\resizebox{0.45\textwidth}{!}{%
\begin{tabular}{l|l}
\toprule
\textbf{Configuration} & \textbf{Value} \\
\midrule
Batch size (per GPU) & 32 \\
Global batch size & 32$\times$64=2048 \\
Action head & Large (1024,24,16)  \\
Action chunk horizon \(H\) & 30 \\
Image resize & 224$\times$224 \\
Action normalization & on \\
Data shuffling & on \\
Optimizer & AdamW \\
Betas & (0.9, 0.95) \\
Weight decay & 0.0 \\
Learning rate & 1e-4 \\
VLM LR multiplier & 0.1 \\
Warm-up steps & 1000 \\
Scheduler & none \\
Training steps & 150K \\
Precision & bfloat16 \\
\bottomrule
\end{tabular}}
\end{table}

\section{Per-Benchmark Ablation Breakdowns}\label{app:per_benchmark_ablations}

The main text (Table~\ref{tab:libero_ablations}) presents detailed ablations on LIBERO. Tables~\ref{tab:calvin_ablations}--\ref{tab:simpler_ablations} below extend the same controlled protocol to CALVIN, WidowX, and Google Robot. In each table, one knob is varied while all others are held at the default configuration. The consistent factor ordering across all three additional benchmarks---training dynamics causing the largest performance swings, followed by task configuration, then architecture---corroborates the attribution hierarchy established in Section~\ref{sec:ablations}.

\paragraph{CALVIN ABC$\rightarrow$D.}
Table~\ref{tab:calvin_ablations} reports ablations under the standard ABC$\rightarrow$D protocol, where the policy is trained on scenes A, B, C and evaluated on 1{,}000 five-step language-conditioned chains in the held-out scene D. CALVIN is particularly sensitive to data pipeline choices: disabling shuffling drops the score from 4.47 to 0.17, and setting the VLM LR multiplier to 1.0 yields a similarly sharp drop to 0.16. The action horizon also plays a notable role ($H{=}10$ drops the score to 3.84), confirming that longer horizons are beneficial for multi-step sequential tasks. \crhl{For this horizon ablation we changed only the action chunk horizon, from the default $H{=}30$ to $H{=}10$.}

\begin{table}[t]
\caption{\textbf{Ablations on CALVIN.} Each row corresponds to one ablation setting with the remaining knobs fixed to the default configuration.}
\label{tab:calvin_ablations}
\centering
\scriptsize
\setlength{\tabcolsep}{3pt}
\resizebox{0.8\textwidth}{!}{%
\begin{tabular}{l|l|cccccc}
\toprule
\textbf{Knob} & \textbf{Value} & \textbf{SR1} & \textbf{SR2} & \textbf{SR3} & \textbf{SR4} & \textbf{SR5} & \textbf{Score}\\
\midrule
\rowcolor{gray!10} \textbf{SimVLA} & Default settings & \textbf{99.4} & \textbf{96.4} & \textbf{89.8} & \textbf{83.7} & \textbf{77.2} & \textbf{4.47}\\
\midrule
\rowcolor{gray!10} \multicolumn{8}{l}{\textit{Data \& representation} \ (default: \(H{=}30\), shuffling on, normalization on)} \\
\midrule
Action chunk horizon \(H\) & \(H=10\) & 96.9 & 86.7 & 76.6 & 66.7 & 56.8 & 3.84 \\
\midrule
Data shuffling & off & 14.7 & 2.2 & 0.4 & 0.1 & 0.0 & 0.17 \\
\midrule
Action normalization & off & 95.4 & 84.7 & 73.7 & 63.5 & 53.7 & 3.73 \\
\midrule
\rowcolor{gray!10} \multicolumn{8}{l}{\textit{Optimization dynamics} \ (default: lr \(1\!\times\!10^{-4}\), warm-up 2000, scheduler cosine, VLM LR multiplier 0.1)} \\
\midrule
\multirow{1}{*}{Learning rate} 
& \(5\!\times\!10^{-5}\) & 98.4 & 92.4 & 83.5 & 71.8 & 63.3 &4.10\\
\midrule
Warm-up steps & 1000 & 97.7 & 91.3 & 82.2 & 71.0 & 62.2 & 4.05\\
\midrule
Scheduler & no cosine & 97.8 & 90.7 & 80.3 & 68.7 & 58.6 & 3.96\\
\midrule
VLM LR multiplier &1.0& 14.8 & 1.4 & 0.0 & 0.0& 0.0& 0.16\\
\midrule
\rowcolor{gray!10} \multicolumn{8}{l}{\textit{Architecture configuration} \ (default: large head (1024,24,16), concat injection)} \\
\midrule
Action transformer scale & small (768,12,12) & 95.7 & 90.3 & 78.1 & 68.1 & 59.2 &3.91\\
\bottomrule
\end{tabular}%
}
\end{table}

\paragraph{WidowX (SimplerEnv-Bridge).}
Table~\ref{tab:widowx_ablations} reports ablations on the four WidowX tabletop tasks evaluated via SimplerEnv. The same hierarchy holds: disabling shuffling drops the average from 95.8\% to 37.5\%, and setting the VLM LR multiplier to 1.0 causes a sharp drop to 11.5\%. Architecture changes are again secondary---switching to a small action head reduces performance to 80.2\%, a much smaller swing than the training-dynamics knobs.

\begin{table}[t]
\caption{\textbf{Ablations on WidowX.} Each row corresponds to one ablation setting with the remaining knobs fixed to the default configuration.}
\label{tab:widowx_ablations}
\centering
\scriptsize
\setlength{\tabcolsep}{3pt}
\resizebox{0.8\textwidth}{!}{%
\begin{tabular}{l|l|ccccc}
\toprule
\textbf{Knob} & \textbf{Value} & \textbf{spoon} & \textbf{carrot} & \textbf{stack} & \textbf{eggplant} & \textbf{Avg}\\
\midrule
\rowcolor{gray!10} \textbf{SimVLA} & Default settings & \textbf{100.0} & \textbf{100.0} & \textbf{91.7} & \textbf{91.7} & \textbf{95.8}\\
\midrule
\rowcolor{gray!10} \multicolumn{7}{l}{\textit{Data \& representation} \ (default: \(H{=}30\), shuffling on, normalization on)} \\
\midrule
Action chunk horizon \(H\) & \(H=10\) & 95.8 & 83.3 & 70.8 & 21.7 & 67.9 \\
\midrule
Data shuffling & off & 54.2 & 50.0 & 29.2 & 16.7 & 37.5 \\
\midrule
Action normalization & off & 79.2 & 95.8 & 79.2 & 34.8 & 72.3 \\
\midrule
\rowcolor{gray!10} \multicolumn{7}{l}{\textit{Optimization dynamics} \ (default: lr \(1\!\times\!10^{-4}\), warm-up 2000, scheduler none, VLM LR multiplier 0.05)} \\
\midrule
Learning rate & \(5\!\times\!10^{-5}\) & 95.8 & 91.7 & 91.7 & 75.0 & 88.6 \\
\midrule
Warm-up steps & 1000 & 95.8 & 95.8 & 91.7 & 75.0 & 89.6 \\
\midrule
Scheduler & cosine & 95.8 & 95.8 & 91.7 & 79.2 & 90.6 \\
\midrule
VLM LR multiplier & 1.0 & 29.2 & 16.7 & 0.0 & 0.0 & 11.5 \\
\midrule
\rowcolor{gray!10} \multicolumn{7}{l}{\textit{Architecture configuration} \ (default: large head (1024,24,16), concat injection)} \\
\midrule
Action transformer scale & small (768,12,12) & 70.8 & 95.8 & 100.0 & 54.2 & 80.2 \\
\bottomrule
\end{tabular}%
}
\end{table}

\paragraph{Google Robot (SimplerEnv-Fractal).}
Table~\ref{tab:simpler_ablations} reports ablations on the Google Robot variant-aggregation tasks (Pick, Move Near, Open/Close Drawer). The action horizon effect is especially pronounced here: reducing $H$ from 30 to 10 drops the average from 77.0\% to 34.2\%, the largest horizon-induced swing across all four benchmarks. Training dynamics remain the largest measured factor---disabling shuffling ($-$24.0 pp) and setting VLM LR multiplier to 1.0 ($-$21.0 pp) cause large degradations---while the small action head is within 1.6 pp of the default.

\begin{table}[h!]
\caption{\textbf{Ablations on Google Robot (Variant Aggregation).} Each row corresponds to one ablation setting with the remaining knobs fixed to the default configuration.}
\label{tab:simpler_ablations}
\centering
\scriptsize
\setlength{\tabcolsep}{3pt}
\resizebox{0.8\textwidth}{!}{%
\begin{tabular}{l|l|cccc}

\toprule
\textbf{Knob} & \textbf{Value} & \textbf{Pick} & \textbf{Move} & \textbf{Open} & \textbf{Avg}\\
\midrule
\rowcolor{gray!10} \textbf{SimVLA} & Default settings & \textbf{82.3} & \textbf{81.0} & \textbf{67.7} & \textbf{77.0}\\
\midrule
\rowcolor{gray!10} \multicolumn{6}{l}{\textit{Data \& representation} \ (default: \(H{=}30\), shuffling on, normalization on)} \\
\midrule
Action chunk horizon \(H\) & \(H=10\) & 29.0 & 32.2 & 41.3 & 34.2 \\
\midrule
Data shuffling & off & 61.2 & 53.8 & 43.9 & 53.0 \\
\midrule
Action normalization & off & 72.5 & 67.8 & 64.8 & 68.4 \\
\midrule
\rowcolor{gray!10} \multicolumn{6}{l}{\textit{Optimization dynamics} \ (default: lr \(1\!\times\!10^{-4}\), warm-up 2000, scheduler none, VLM LR multiplier 0.01)} \\
\midrule
Learning rate & \(5\!\times\!10^{-5}\) & 81.3 & 73.3 & 67.7 & 74.1 \\
\midrule
Warm-up steps & 1000 & 75.0 & 78.2 & 66.1 & 73.1 \\
\midrule
Scheduler & cosine & 79.8 & 77.0 & 67.9 & 74.9 \\
\midrule
VLM LR multiplier & 1.0 & 62.8 & 50.3 & 54.8 & 56.0 \\
\midrule
\rowcolor{gray!10} \multicolumn{6}{l}{\textit{Architecture configuration} \ (default: large head (1024,24,16), concat injection)} \\
\midrule
Action transformer scale & small (768,12,12) & 81.9 & 78.3 & 66.1 & 75.4 \\
\bottomrule
\end{tabular}%
}

\end{table}

\section{Qualitative Rollouts}\label{app:qualitative_rollouts}

Figures~\ref{fig:qual_libero}--\ref{fig:qual_google_robot} show representative rollout sequences from three simulation benchmarks, illustrating the diversity of manipulation skills that SimVLA executes successfully under its default configuration. Each figure displays keyframes from a successful episode, demonstrating that SimVLA produces smooth, temporally consistent action trajectories across diverse manipulation scenarios.

\section{BEHAVIOR-1K Challenge Results}\label{app:behavior1k}

To assess whether SimVLA's competence extends beyond fixed-task simulation benchmarks, we conduct a preliminary evaluation on the BEHAVIOR-1K Challenge~\citep{BEHAVIOR1K}. BEHAVIOR-1K tests open-world robustness: the robot operates in a full household environment with randomized object layouts, initial poses, and scene configurations. Table~\ref{tab:b1k_eval} reports results on Task~1 (\emph{Turning on Radio}), where the agent must navigate to a radio, interact with it, and successfully turn it on under randomized conditions. SimVLA achieves a task success score of 0.40, outperforming $\pi_{0.5}$ (3B backbone, pretrained on 200k+ robot episodes), which scores 0.00, and approaching the 1st-place challenge winner (0.60)~\citep{larchenko2025behavior}. This result is notable because SimVLA uses only a small backbone and no additional large-scale robot-data VLA pretraining before benchmark-specific training, yet remains competitive in an open-world setting designed to stress-test generalization beyond the controlled environments of LIBERO or CALVIN. SimVLA also succeeds on Task~2 (\emph{Picking Up Trash})---a 3--5 minute long-horizon episode requiring the robot to navigate from a random starting pose, search for scattered trash, grasp each piece, and place it in a bin. Training and inference settings are reported in Table~\ref{tab:hparams_libero}.

\begin{table}[t]
\centering\crhlon
\caption{\crhl{\textbf{BEHAVIOR-1K Challenge} task success scores under the challenge protocol (open-world, randomized poses and objects).}}
\label{tab:b1k_eval}
\scriptsize
\setlength{\tabcolsep}{4pt}
\begin{tabular}{l|ccc}
\toprule
Task & \textbf{SimVLA} & $\pi_{0.5}$ & 1st-place winner~\citep{larchenko2025behavior} \\
\midrule
Task 1: Turning on Radio & \textbf{0.40} & 0.00 & 0.60 \\
Task 2: Picking Up Trash & \textbf{0.50} & 0.00 & 1.00 \\
\bottomrule
\end{tabular}
\end{table}



\begin{figure}[h!]
\centering
\includegraphics[width=\textwidth]{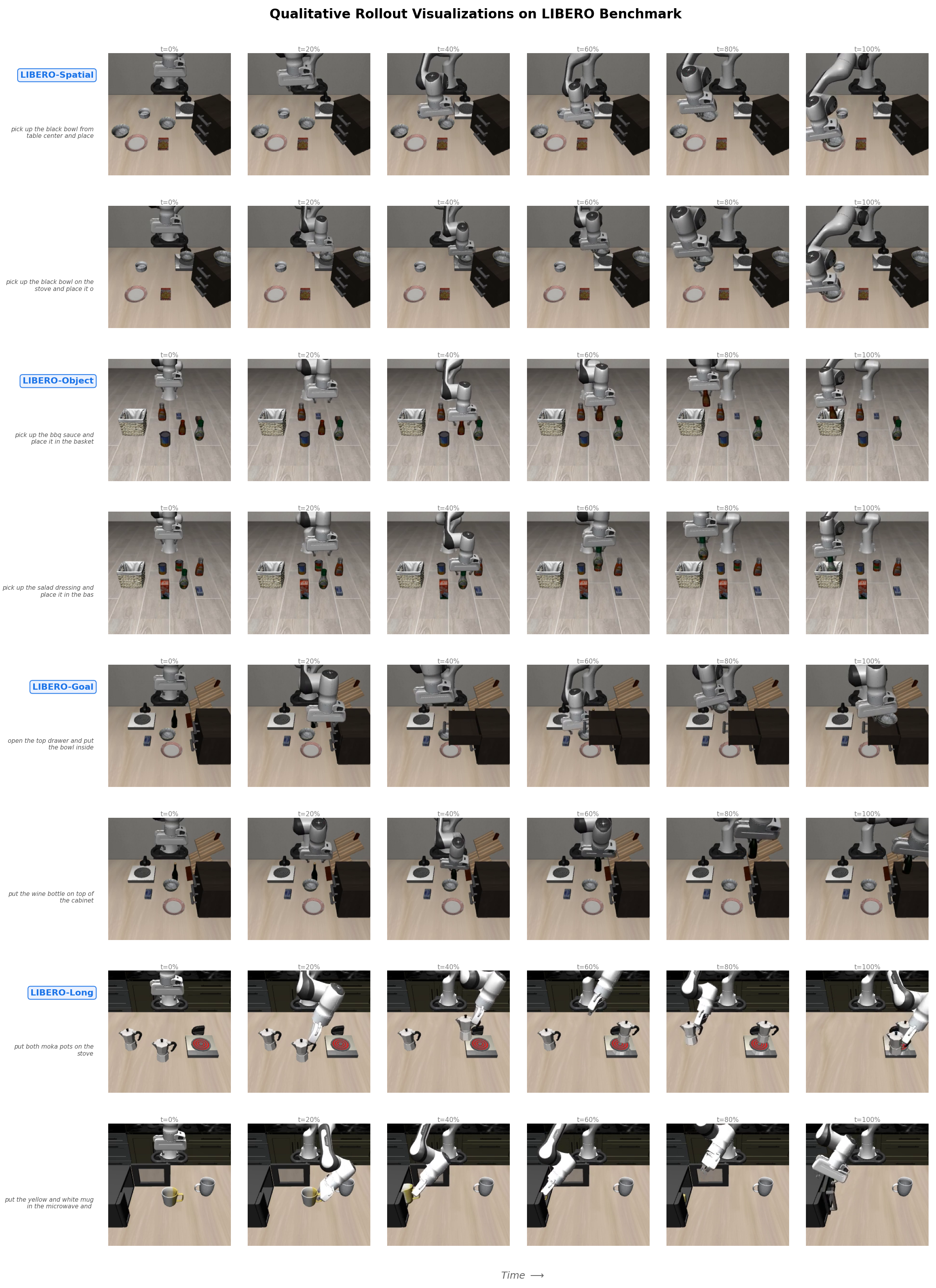}
\caption{\textbf{Qualitative Rollout on LIBERO.} Keyframes from a representative long-horizon kitchen manipulation task. The policy must execute a sequence of precise pick-and-place actions in the correct order. SimVLA completes the full multi-step sequence with smooth transitions between subtasks, reflecting strong temporal consistency under the $H{=}10$ action chunking horizon used for LIBERO.}
\label{fig:qual_libero}
\end{figure}

\begin{figure}[h!]
\centering
\includegraphics[width=\textwidth]{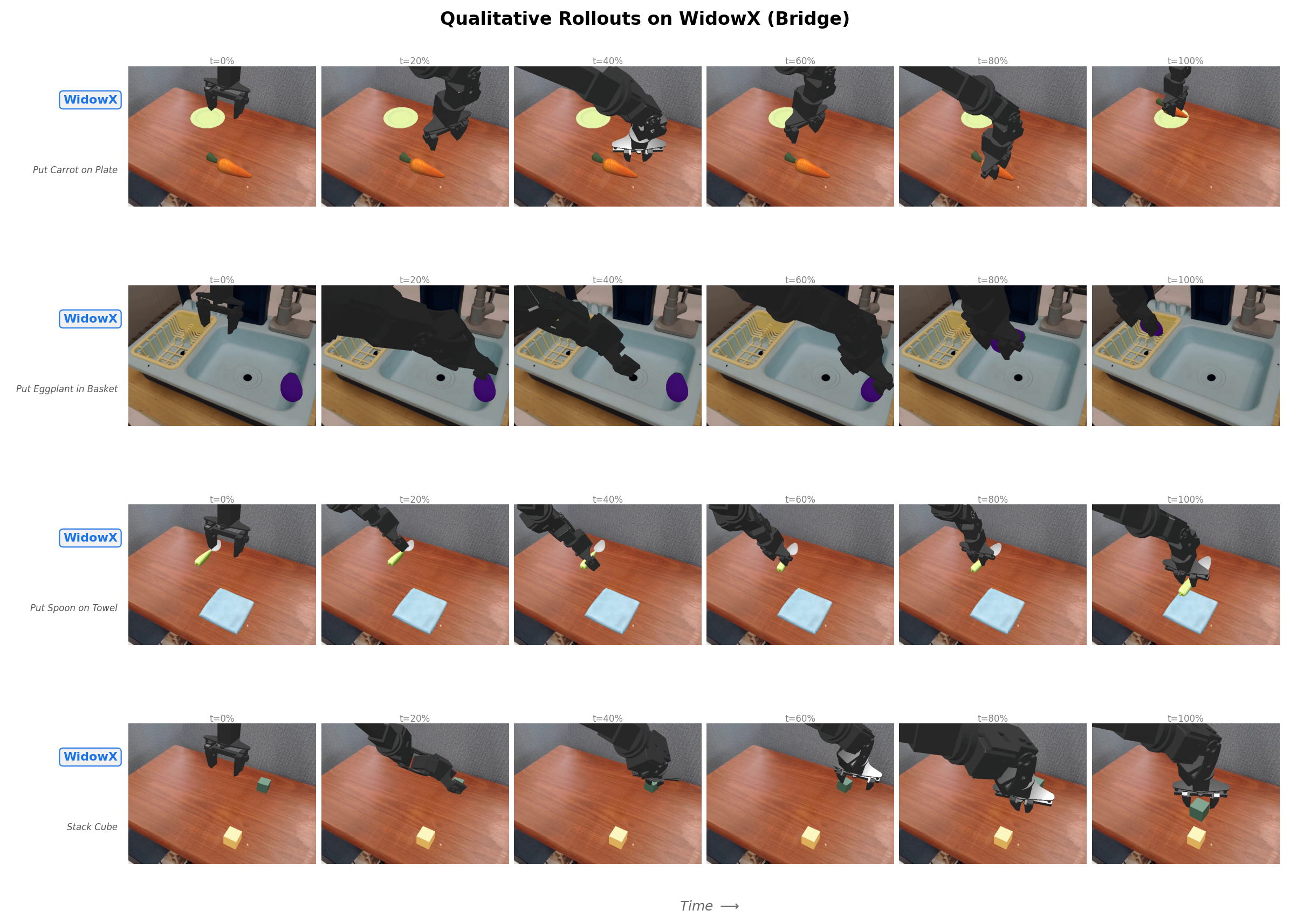}
\caption{\textbf{Qualitative Rollout on WidowX.} Keyframes from a tabletop pick-and-place task in the SimplerEnv-Bridge environment. The WidowX robot must locate, grasp, and transport an object to a target location from a single third-person RGB view. SimVLA achieves 95.8\% average success on this benchmark, matching X-VLA without additional large-scale robot-data VLA pretraining before benchmark-specific training.}
\label{fig:qual_widowx}
\end{figure}

\begin{figure}[h!]
\centering
\includegraphics[width=\textwidth]{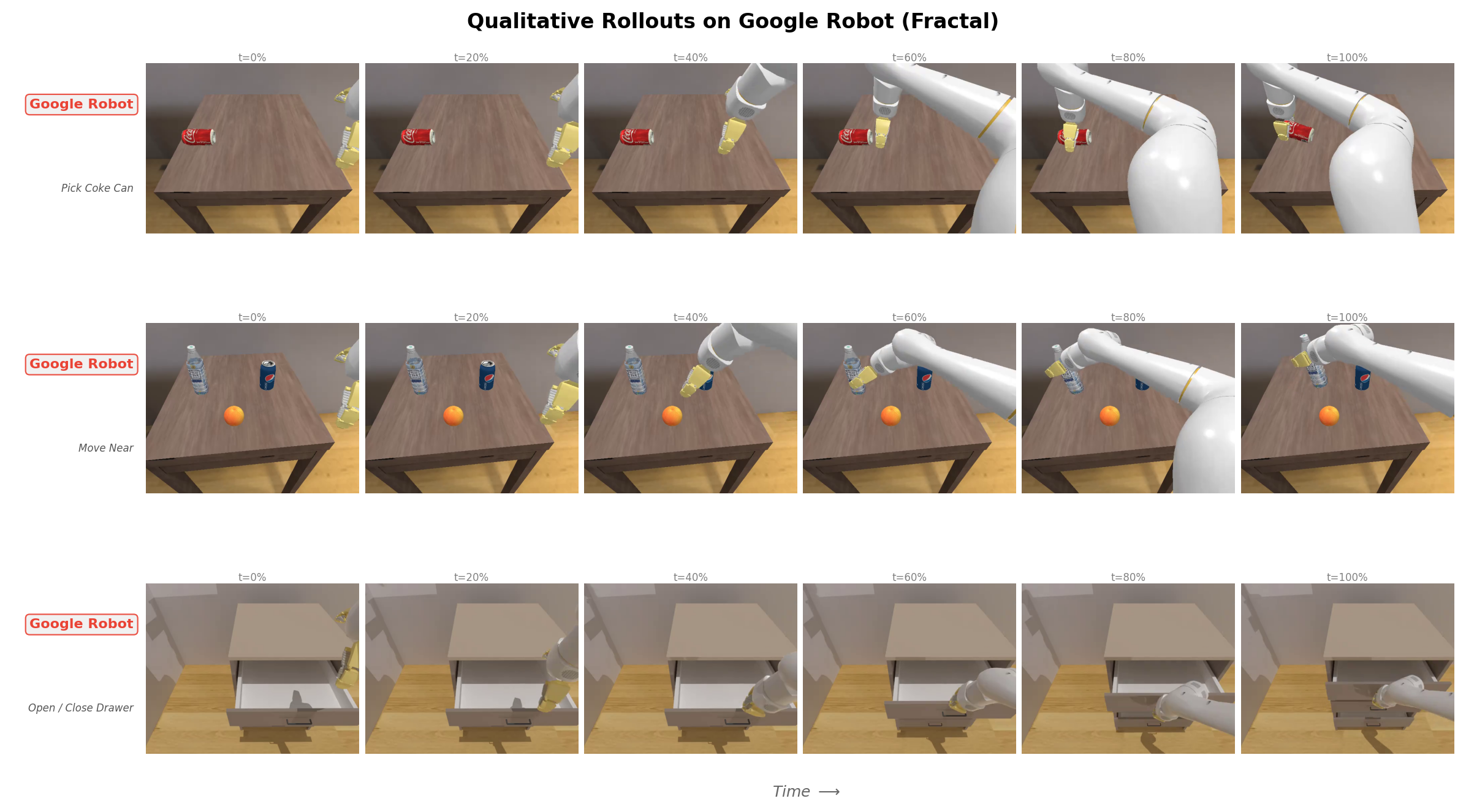}
\caption{\textbf{Qualitative Rollout on Google Robot.} Keyframes from a drawer manipulation task in the SimplerEnv-Fractal environment. The policy controls a mobile manipulator to approach and operate a drawer handle. SimVLA achieves 77.0\% variant-aggregation average across Pick, Move Near, and Open/Close Drawer tasks on this benchmark.}
\label{fig:qual_google_robot}
\end{figure}

\clearpage
{\crhlon
\section{Robustness of the Attribution Hierarchy}\label{app:robustness}

\begin{table}[t]
\centering\crhlon
\caption{\crhl{\textbf{LIBERO one-knob-at-a-time sweep at the small-head base} (\{768,12,12\}). Success rate (\%).}}
\label{tab:app_libero_small}
\scriptsize
\setlength{\tabcolsep}{3pt}
\resizebox{0.7\textwidth}{!}{%
\begin{tabular}{l|l|ccccc}
\toprule
\textbf{Knob} & \textbf{Value} & \textbf{Spatial} & \textbf{Object} & \textbf{Goal} & \textbf{Long} & \textbf{Avg} \\
\midrule
\rowcolor{gray!10} SimVLA & Default & 98.8 & 99.6 & 98.6 & 94.8 & 98.0 \\
\midrule
Horizon & $H{=}20$ & 96.4 & 96.6 & 93.4 & 79.8 & 91.6 \\
 & $H{=}30$ & 97.4 & 99.2 & 78.6 & 75.4 & 87.7 \\
\midrule
Shuffling & off & 10.8 & 0.0 & 14.2 & 0.0 & 6.3 \\
\midrule
Norm & off & 13.4 & 8.6 & 23.8 & 0.0 & 11.4 \\
\midrule
LR & $5\!\times\!10^{-5}$ & 99.4 & 99.4 & 95.4 & 70.4 & 91.2 \\
 & $1\!\times\!10^{-4}$ & 98.8 & 99.6 & 98.6 & 94.2 & 97.8 \\
 & $5\!\times\!10^{-4}$ & 89.0 & 99.0 & 59.0 & 64.0 & 77.8 \\
\midrule
Warm-up & 1000 & 99.4 & 99.0 & 98.0 & 90.8 & 96.8 \\
\midrule
Scheduler & cosine & 99.0 & 98.2 & 97.2 & 91.6 & 96.5 \\
\midrule
VLM LR mult & 1.0 & 64.4 & 78.2 & 73.8 & 13.6 & 57.5 \\
\midrule
Injection & AdaLN & 98.2 & 99.2 & 96.7 & 80.0 & 93.5 \\
 & cross-attn & 98.2 & 98.6 & 92.4 & 82.4 & 92.9 \\
\midrule
Backbone & Florence-2 & 99.4 & 99.0 & 98.0 & 90.8 & 96.8 \\
\bottomrule
\end{tabular}%
}
\end{table}

\begin{table}[t]
\centering\crhlon
\caption{\crhl{\textbf{WidowX one-knob-at-a-time sweep at the small-head base.} Success rate (\%). }}
\label{tab:app_widowx_small}
\scriptsize
\setlength{\tabcolsep}{3pt}
\resizebox{0.8\textwidth}{!}{%
\begin{tabular}{l|l|ccccc}
\toprule
\textbf{Knob} & \textbf{Value} & \textbf{Spoon} & \textbf{Carrot} & \textbf{Stack} & \textbf{Eggplant} & \textbf{Avg} \\
\midrule
\rowcolor{gray!10} SimVLA & Default & 70.8 & 95.8 & 100.0 & 54.2 & 80.2 \\
\midrule
Horizon & $H{=}10$ & 70.8 & 79.2 & 83.3 & 16.7 & 62.5 \\
\midrule
Shuffling & off & 41.7 & 50.0 & 33.3 & 0.0 & 31.3 \\
\midrule
Norm & off & 66.7 & 95.8 & 83.3 & 0.0 & 61.5 \\
\midrule
LR & $5\!\times\!10^{-5}$ & 95.8 & 95.8 & 100.0 & 58.3 & 87.5 \\
\midrule
Warm-up & 1000 & 91.7 & 100.0 & 87.5 & 20.8 & 75.0 \\
\midrule
Scheduler & cosine & 83.3 & 95.8 & 95.8 & 41.7 & 79.2 \\
\midrule
VLM LR mult & 1.0 & 16.7 & 16.7 & 4.2 & 0.0 & 9.4 \\
\bottomrule
\end{tabular}%
}
\end{table}

\begin{table}[t]
\centering\crhlon
\caption{\crhl{\textbf{CALVIN ABC$\rightarrow$D one-knob-at-a-time sweep at the small-head base.} ``Scheduler: no cosine'' removes the default cosine decay.}}
\label{tab:app_calvin_small}
\scriptsize
\setlength{\tabcolsep}{3pt}
\resizebox{0.8\textwidth}{!}{%
\begin{tabular}{l|l|cccccc}
\toprule
\textbf{Knob} & \textbf{Value} & \textbf{SR1} & \textbf{SR2} & \textbf{SR3} & \textbf{SR4} & \textbf{SR5} & \textbf{Avg Length} \\
\midrule
\rowcolor{gray!10} SimVLA & Default & 95.7 & 90.3 & 78.1 & 68.1 & 59.2 & 3.91 \\
\midrule
Horizon & $H{=}10$ & 88.5 & 79.0 & 69.8 & 60.9 & 51.8 & 3.50 \\
\midrule
Shuffling & off & 18.6 & 2.1 & 0.3 & 0.0 & 0.0 & 0.21 \\
\midrule
Norm & off & 89.0 & 79.0 & 68.7 & 59.2 & 50.1 & 3.46 \\
\midrule
LR & $5\!\times\!10^{-5}$ & 99.4 & 95.8 & 90.2 & 85.0 & 76.6 & 4.47 \\
\midrule
Warm-up & 1000 & 99.0 & 94.4 & 86.6 & 79.2 & 68.0 & 4.27 \\
\midrule
Scheduler & no cosine & 98.6 & 95.8 & 91.0 & 85.2 & 75.8 & 4.46 \\
\midrule
VLM LR mult & 1.0 & 16.9 & 1.1 & 0.0 & 0.0 & 0.0 & 0.18 \\
\bottomrule
\end{tabular}%
}
\end{table}

\begin{table}[t]
\centering\crhlon
\caption{\crhl{\textbf{LIBERO one-knob-at-a-time sweep at the Florence-2 (0.9B) base} with the large action head. Success rate (\%).}}
\label{tab:app_libero_florence}
\scriptsize
\setlength{\tabcolsep}{3pt}
\resizebox{0.8\textwidth}{!}{%
\begin{tabular}{l|l|ccccc}
\toprule
\textbf{Knob} & \textbf{Value} & \textbf{Spatial} & \textbf{Object} & \textbf{Goal} & \textbf{Long} & \textbf{Avg} \\
\midrule
\rowcolor{gray!10} SimVLA & Default & 99.8 & 99.2 & 98.0 & 93.8 & 97.7 \\
\midrule
Horizon & $H{=}20$ & 97.4 & 95.4 & 90.6 & 87.2 & 92.6 \\
 & $H{=}30$ & 97.2 & 91.6 & 88.2 & 68.8 & 86.5 \\
\midrule
Shuffling & off & 7.8 & 0.0 & 26.4 & 0.0 & 8.6 \\
\midrule
Norm & off & 20.8 & 0.0 & 26.4 & 0.0 & 11.8 \\
\midrule
LR & $5\!\times\!10^{-5}$ & 95.1 & 99.2 & 93.4 & 82.6 & 92.6 \\
 & $1\!\times\!10^{-4}$ & 99.0 & 98.2 & 97.2 & 91.6 & 96.5 \\
 & $5\!\times\!10^{-4}$ & 95.8 & 84.6 & 84.4 & 49.2 & 78.5 \\
\midrule
Warm-up & 1000 & 99.4 & 98.0 & 97.0 & 90.2 & 96.1 \\
\midrule
Scheduler & cosine & 99.8 & 99.2 & 98.0 & 93.0 & 97.5 \\
\midrule
VLM LR mult & 1.0 & 76.4 & 99.2 & 13.8 & 13.4 & 50.7 \\
\midrule
Injection & AdaLN & 98.6 & 96.0 & 92.2 & 74.4 & 90.3 \\
 & cross-attn & 98.6 & 96.2 & 96.0 & 72.4 & 90.8 \\
\midrule
Action head & Small & 98.2 & 98.4 & 98.4 & 94.6 & 97.4 \\
\bottomrule
\end{tabular}%
}
\end{table}

\begin{table}[t]
\centering\crhlon
\caption{\crhl{\textbf{Autoregressive-head experiment on LIBERO.} The continuous flow-matching head is replaced by a discrete autoregressive action head (256 bins) on the same SmolVLM-0.5B encoder; the AR default uses cosine scheduling. Success rate (\%).}}
\label{tab:app_libero_ar}
\scriptsize
\setlength{\tabcolsep}{3pt}
\resizebox{0.8\textwidth}{!}{%
\begin{tabular}{l|l|ccccc}
\toprule
\textbf{Knob} & \textbf{Value} & \textbf{Spatial} & \textbf{Object} & \textbf{Goal} & \textbf{Long} & \textbf{Avg} \\
\midrule
\rowcolor{gray!10} SimVLA (AR) & Default & 94.8 & 99.4 & 92.6 & 94.8 & 95.4 \\
\midrule
Horizon & $H{=}20$ & 98.2 & 95.6 & 93.8 & 82.0 & 92.4 \\
 & $H{=}30$ & 96.0 & 97.0 & 81.6 & 87.0 & 90.4 \\
\midrule
Shuffling & off & 0.0 & 0.0 & 0.0 & 0.0 & 0.0 \\
\midrule
LR & $5\!\times\!10^{-5}$ & 97.4 & 96.2 & 91.6 & 81.8 & 91.8 \\
 & $1\!\times\!10^{-4}$ & 97.2 & 95.2 & 91.6 & 92.2 & 94.0 \\
 & $5\!\times\!10^{-4}$ & 0.0 & 0.0 & 0.0 & 0.0 & 0.0 \\
\midrule
Warm-up & 1000 & 97.8 & 99.8 & 90.0 & 86.6 & 93.5 \\
\midrule
Scheduler & no cosine & 94.4 & 93.0 & 65.8 & 63.8 & 79.2 \\
\midrule
VLM LR multiplier & 1.0 & 39.8 & 60.4 & 47.2 & 0.0 & 36.9 \\
\midrule
Head scale & small & 94.6 & 98.2 & 88.0 & 89.0 & 92.5 \\
\bottomrule
\end{tabular}%
}
\end{table}

\begin{table}[htbp]
\centering\crhlon
\small
\setlength{\tabcolsep}{3pt}

\caption{\crhl{\textbf{LIBERO, three training seeds.} X-VLA is reproduced with the same three-seed procedure.}}
\label{tab:app_seed_libero}
\begin{tabular}{l|ccccc}
\toprule
\textbf{Method} & \textbf{Spatial} & \textbf{Object} & \textbf{Goal} & \textbf{Long} & \textbf{Average} \\
\midrule
\rowcolor{gray!10} SimVLA & \textbf{99.4 $\pm$ 0.5} & \textbf{99.6 $\pm$ 0.4} & \textbf{98.0 $\pm$ 0.6} & \textbf{95.4 $\pm$ 0.8} & \textbf{98.1 $\pm$ 0.5} \\
XVLA & 98.2 $\pm$ 0.8 & 98.5 $\pm$ 0.5 & 97.2 $\pm$ 0.4 & 96.3 $\pm$ 0.5 & 97.5 $\pm$ 0.4 \\
\bottomrule
\end{tabular}

\vspace{6pt}

\caption{\crhl{\textbf{WidowX, three training seeds.}}}
\label{tab:app_seed_widowx}
\begin{tabular}{l|ccccc}
\toprule
\textbf{Method} & \textbf{Carrot on Plate} & \textbf{Eggplant in Basket} & \textbf{Spoon on Towel} & \textbf{Stack Cube} & \textbf{Average} \\
\midrule
\rowcolor{gray!10} SimVLA & \textbf{97.2 $\pm$ 2.4} & \textbf{86.1 $\pm$ 9.6} & \textbf{98.6 $\pm$ 2.4} & \textbf{93.1 $\pm$ 2.4} & \textbf{93.8 $\pm$ 2.1} \\
\bottomrule
\end{tabular}

\vspace{6pt}

\caption{\crhl{\textbf{Google Robot, three training seeds.}}}
\label{tab:app_seed_google}
\begin{tabular}{l|cccc}
\toprule
\textbf{Method} & \textbf{Coke Can} & \textbf{Move Near} & \textbf{Open/Close} & \textbf{VA Average} \\
\midrule
\rowcolor{gray!10} SimVLA & \textbf{83.5 $\pm$ 1.6} & \textbf{77.1 $\pm$ 3.4} & \textbf{68.2 $\pm$ 0.5} & \textbf{76.3 $\pm$ 0.8} \\
\bottomrule
\end{tabular}

\vspace{6pt}

\caption{\crhl{\textbf{CALVIN ABC$\rightarrow$D, three training seeds.}}}
\label{tab:app_seed_calvin}
\begin{tabular}{l|cccccc}
\toprule
\textbf{Method} & \textbf{Average Length} & \textbf{SR@1 (\%)} & \textbf{SR@2 (\%)} & \textbf{SR@3 (\%)} & \textbf{SR@4 (\%)} & \textbf{SR@5 (\%)} \\
\midrule
\rowcolor{gray!10} SimVLA & \textbf{4.4 $\pm$ 0.1} & \textbf{99.3 $\pm$ 0.5} & \textbf{96.1 $\pm$ 0.4} & \textbf{89.1 $\pm$ 0.7} & \textbf{82.9 $\pm$ 0.8} & \textbf{73.7 $\pm$ 3.1} \\
\bottomrule
\end{tabular}
\end{table}

\begin{table}[t]
\centering\crhlon
\caption{\crhl{\textbf{Table~\ref{tab:libero_ablations} with three training seeds (LIBERO).} Mean $\pm$ sample standard deviation over seeds 0, 1, 2. The default row is the three-seed mean of the SimVLA default.}}
\label{tab:app_libero_3seed}
\scriptsize
\setlength{\tabcolsep}{3pt}
\resizebox{0.9\textwidth}{!}{%
\begin{tabular}{l|l|ccccc}
\toprule
\textbf{Knob} & \textbf{Value} & \textbf{Spatial} & \textbf{Object} & \textbf{Goal} & \textbf{Long} & \textbf{Avg} \\
\midrule
\rowcolor{gray!10} SimVLA & Default & 99.4 $\pm$ 0.5 & 99.6 $\pm$ 0.4 & 98.0 $\pm$ 0.6 & 95.4 $\pm$ 0.8 & 98.1 $\pm$ 0.5 \\
\midrule
Horizon & $H{=}20$ & 99.4 $\pm$ 0.7 & 88.0 $\pm$ 1.3 & 92.0 $\pm$ 1.4 & 87.8 $\pm$ 2.1 & 91.8 $\pm$ 1.0 \\
 & $H{=}30$ & 95.8 $\pm$ 1.8 & 92.6 $\pm$ 0.9 & 81.4 $\pm$ 2.6 & 77.8 $\pm$ 4.2 & 86.9 $\pm$ 1.6 \\
\midrule
Shuffling & off & 5.4 $\pm$ 5.1 & 0.0 $\pm$ 0.0 & 12.2 $\pm$ 7.3 & 0.0 $\pm$ 0.0 & 4.4 $\pm$ 0.0 \\
\midrule
Norm & off & 21.4 $\pm$ 6.8 & 2.6 $\pm$ 2.9 & 24.0 $\pm$ 5.4 & 0.0 $\pm$ 0.0 & 12.0 $\pm$ 3.2 \\
\midrule
LR & $5\!\times\!10^{-5}$ & 97.6 $\pm$ 0.6 & 98.2 $\pm$ 1.1 & 95.4 $\pm$ 0.8 & 69.6 $\pm$ 3.7 & 90.2 $\pm$ 1.2 \\
 & $1\!\times\!10^{-4}$ & 99.2 $\pm$ 0.4 & 97.6 $\pm$ 0.7 & 98.0 $\pm$ 0.5 & 84.4 $\pm$ 2.6 & 94.8 $\pm$ 1.0 \\
 & $5\!\times\!10^{-4}$ & 83.8 $\pm$ 3.5 & 92.4 $\pm$ 1.6 & 75.2 $\pm$ 4.8 & 37.8 $\pm$ 7.1 & 72.3 $\pm$ 3.4 \\
\midrule
Warm-up & 1000 & 97.0 $\pm$ 1.2 & 99.4 $\pm$ 0.3 & 96.8 $\pm$ 0.9 & 92.8 $\pm$ 1.5 & 96.5 $\pm$ 0.7 \\
\midrule
Scheduler & cosine & 99.0 $\pm$ 0.4 & 99.0 $\pm$ 0.6 & 98.0 $\pm$ 0.5 & 92.0 $\pm$ 1.3 & 97.0 $\pm$ 0.6 \\
\midrule
VLM LR mult & 1.0 & 40.4 $\pm$ 8.3 & 81.2 $\pm$ 3.2 & 45.8 $\pm$ 6.9 & 7.8 $\pm$ 3.6 & 43.8 $\pm$ 4.1 \\
\midrule
Head scale & small & 98.4 $\pm$ 0.7 & 99.2 $\pm$ 0.3 & 98.0 $\pm$ 0.4 & 93.6 $\pm$ 1.6 & 97.3 $\pm$ 0.6 \\
\midrule
Injection & AdaLN & 99.0 $\pm$ 0.5 & 97.6 $\pm$ 0.9 & 96.2 $\pm$ 0.7 & 69.2 $\pm$ 0.9 & 90.5 $\pm$ 0.8 \\
 & cross-attn & 98.6 $\pm$ 0.6 & 96.2 $\pm$ 1.1 & 95.2 $\pm$ 1.0 & 75.2 $\pm$ 1.3 & 91.3 $\pm$ 0.9 \\
\midrule
Backbone & Florence-2 & 99.4 $\pm$ 0.3 & 99.0 $\pm$ 0.5 & 97.6 $\pm$ 0.4 & 92.0 $\pm$ 1.1 & 97.0 $\pm$ 0.5 \\
\midrule
Objective & DDPM & 96.2 $\pm$ 1.4 & 99.0 $\pm$ 0.4 & 96.4 $\pm$ 0.9 & 89.2 $\pm$ 1.7 & 95.2 $\pm$ 0.8 \\
 & MSE & 96.6 $\pm$ 0.8 & 95.8 $\pm$ 1.5 & 96.0 $\pm$ 0.6 & 89.6 $\pm$ 2.0 & 94.5 $\pm$ 1.1 \\
\bottomrule
\end{tabular}%
}
\end{table}

\begin{table}[t]
\centering\crhlon
\caption{\crhl{\textbf{Backbone families on LIBERO (success rate, \%).} Qwen3-VL-2B is a third backbone family; $\Delta$ is the drop from the SmolVLM default.}}
\label{tab:app_backbones}
\scriptsize
\setlength{\tabcolsep}{3pt}
\resizebox{0.85\textwidth}{!}{%
\begin{tabular}{l|c|ccccc|c}
\toprule
\textbf{Backbone} & \textbf{Params} & \textbf{Spatial} & \textbf{Object} & \textbf{Goal} & \textbf{Long} & \textbf{Avg} & \textbf{$\Delta$ vs default} \\
\midrule
\rowcolor{gray!10} \textbf{SmolVLM} (default) & 0.5B & 99.6 & 99.8 & 98.6 & 96.4 & \textbf{98.6} & — \\
\textbf{Florence-2} & 0.9B & 99.8 & 99.2 & 98.0 & 93.8 & \textbf{97.7} & 0.9 \\
\textbf{Qwen3-VL-2B} & 2B & 99.0 & 98.2 & 96.6 & 94.4 & \textbf{97.1} & 1.5 \\
\bottomrule
\end{tabular}%
}
\end{table}

This appendix provides additional experiments that support the robustness of the attribution hierarchy of Section~\ref{sec:ablations} (training dynamics $>$ task configuration $>$ architecture). Because the main-text ablations vary one knob at a time around a single default, we examine whether the ordering (i) persists when the base configuration changes (Section~\ref{app:robust_bases}), (ii) extends to a different action-generation family (Section~\ref{app:robust_ar}), (iii) is large relative to seed-to-seed variation (Section~\ref{app:robust_seeds}), and (iv) is stable when a third backbone family is added (Section~\ref{app:robust_backbone}). Unless stated otherwise, all runs follow the protocol of Section~\ref{sec:training}: one knob is changed at a time and every other setting stays at the reference value of the respective base configuration. We call the reference configuration of the main text the \emph{default base}.

\subsection{Sweeps around alternative base configurations}\label{app:robust_bases}
The main-text ablations vary one knob around a single default. To test whether the ordering is a property of that default point, we repeat the sweep around further bases:
\begin{itemize}[leftmargin=*,noitemsep,topsep=0pt]
  \item \textbf{Small-head base on three benchmarks:}  small action head $\{768,12,12\}$ on LIBERO (Table~\ref{tab:app_libero_small}), WidowX (Table~\ref{tab:app_widowx_small}) and CALVIN (Table~\ref{tab:app_calvin_small}).
  \item \textbf{Florence-2 base on LIBERO:} the Florence-2 backbone (Table~\ref{tab:app_libero_florence}).
\end{itemize}
At every base the largest training-dynamics drop exceeds the largest task-configuration drop, which in turn exceeds the largest architecture drop wherever architecture knobs were swept. The shuffling and horizon deltas are nearly identical across the three LIBERO base configurations (SmolVLM + small head, Florence-2 + large head, plus the SmolVLM + large head default), while the VLM LR multiplier delta varies more ($54.4$, $40.5$, $47.0$) but stays far above every task-configuration and architecture delta. The training-dynamics maximum stays at roughly $89$--$94$ points and the task-configuration maximum at roughly $10$--$11$ points across the three LIBERO bases, so the two tiers remain separated by about an order of magnitude at each base. Extending to CALVIN and WidowX at the small head base, the training-dynamics vs. task-configuration ratio of tier-max deltas is 9.1 and 4.0 respectively. Neither the tier ordering flips nor the tier-max ranges overlap on any of the configurations tested.

\paragraph{Head-size effect after matched tuning.}
In the main-text ablations, the small-head row on each benchmark reuses the default hyperparameters of the large head (Tables~\ref{tab:libero_ablations}, \ref{tab:calvin_ablations}--\ref{tab:simpler_ablations}). The small-head sweeps above show that part of the larger head-size gaps on CALVIN and WidowX reflects this per-benchmark tuning rather than head capacity. 

\subsection{An autoregressive action head}\label{app:robust_ar}
To test a family outside the continuous flow-matching head, we replace the head with a discrete autoregressive action head with 256 bins on top of the same SmolVLM-0.5B encoder and run the same representative sweep on LIBERO (Table~\ref{tab:app_libero_ar}). The AR default uses cosine scheduling and reaches $95.4\%$ average success, $3.2$ points below the continuous-head default. The tier ordering is preserved: training dynamics (shuffling off or a learning rate of $5\!\times\!10^{-4}$: $95.4\rightarrow 0.0$; VLM LR multiplier $1.0$: $95.4\rightarrow 36.9$), then task configuration (largest drop $5.0$ points, $H{=}30$), then architecture (largest drop $2.9$ points, small head). This experiment covers a single benchmark and a single encoder, so whether the hierarchy transfers to autoregressive-tokenization VLAs in general remains open.

\subsection{Seed variance}\label{app:robust_seeds}
All numbers in the main-text tables were obtained from a single training run (seed~0) and list the best checkpoint of that run. We retrained the default SimVLA with three independent seeds (0, 1, 2) on every benchmark and report mean $\pm$ sample standard deviation. As expected, the three-seed means are slightly lower than the single best-run numbers (for instance LIBERO $98.6\rightarrow 98.1\pm 0.5$). For a comparison under the same protocol we also reproduced X-VLA~\citep{x_VLA} with three seeds on LIBERO; its average is $97.5\pm 0.4$ (Table~\ref{tab:app_seed_libero}), versus the $98.1$ reported in the original paper.

Table~\ref{tab:app_libero_3seed} repeats the LIBERO ablation table with three seeds for every row. Using the default standard deviation ($\pm 0.5$) as the noise scale, the head-scale delta ($0.8$) and the Florence-2 backbone delta ($1.1$) are within the seed-noise scale, whereas the injection-scheme deltas ($6.8$--$7.6$) and the action-objective deltas ($2.9$--$3.6$) are clearly above it. The training-dynamics deltas (shuffling off: $93.7$ points, VLM LR multiplier $1.0$: $54.3$ points) are one to two orders of magnitude larger than the seed standard deviations. The architecture tier therefore remains the smallest measured tier whether its small deltas are treated as real or as noise. 



\subsection{A third backbone family}\label{app:robust_backbone}
The backbone row of Table~\ref{tab:libero_ablations} swaps SmolVLM-0.5B for Florence-2-0.9B and therefore changes size, pretraining corpus, model family and training objective at once. It should be read as a backbone-family swap and not as a controlled scale ablation. We additionally evaluated Qwen3-VL-2B~\citep{qwen3vl} under the default LIBERO configuration (Table~\ref{tab:app_backbones}). Over a $4\times$ range in backbone size ($0.5$B$\rightarrow 0.9$B$\rightarrow 2$B), the backbone-swap deltas stay at $0.9$ (Florence-2) and $1.5$ (Qwen3-VL-2B) points, much smaller than the training-dynamics maximum.

}

\clearpage
\section*{NeurIPS Paper Checklist}

\begin{enumerate}

\item {\bf Claims}
    \item[] Question: Do the main claims made in the abstract and introduction accurately reflect the paper's contributions and scope?
    \item[] Answer: \answerYes{}
    \item[] Justification: The abstract and introduction state the attribution hierarchy with quantitative ranges, \crhl{report benchmark results (single-run and three-seed), describe the contribution as methodological and empirical, and explicitly} scope the findings to the simple VLM-encoder plus lightweight-head family (Section~\ref{sec:conc}, Limitations).

\item {\bf Limitations}
    \item[] Question: Does the paper discuss the limitations of the work performed by the authors?
    \item[] Answer: \answerYes{}
    \item[] Justification: A dedicated Limitations subsection in Section~\ref{sec:conc} discusses limitations.

\item {\bf Theory assumptions and proofs}
    \item[] Question: For each theoretical result, does the paper provide the full set of assumptions and a complete (and correct) proof?
    \item[] Answer: \answerNA{}
    \item[] Justification: This paper is empirical and does not include theoretical results or proofs.

    \item {\bf Experimental result reproducibility}
    \item[] Question: Does the paper fully disclose all the information needed to reproduce the main experimental results of the paper to the extent that it affects the main claims and/or conclusions of the paper (regardless of whether the code and data are provided or not)?
    \item[] Answer: \answerYes{}
    \item[] Justification: Section~\ref{sec:training} details the training recipe (normalization, shuffling, optimization). The Appendix (Training Hyperparameters) provides full hyperparameter tables for all benchmarks. Code is available at \url{https://frontierrobo.github.io/SimVLA/}.

\item {\bf Open access to data and code}
    \item[] Question: Does the paper provide open access to the data and code, with sufficient instructions to faithfully reproduce the main experimental results, as described in supplemental material?
    \item[] Answer: \answerYes{}
    \item[] Justification: Code is available at \url{https://frontierrobo.github.io/SimVLA/}. All simulation benchmarks (LIBERO, CALVIN, WidowX, Google Robot) use publicly available datasets. The Galaxea Open-World Dataset is described in the Appendix.

\item {\bf Experimental setting/details}
    \item[] Question: Does the paper specify all the training and test details (e.g., data splits, hyperparameters, how they were chosen, type of optimizer) necessary to understand the results?
    \item[] Answer: \answerYes{}
    \item[] Justification: Training details are in Section~\ref{sec:training} and the Appendix (Training Hyperparameters). Evaluation protocols follow each benchmark's standard setup as described in Section~\ref{sec:evaluation}.

\item {\bf Experiment statistical significance}
    \item[] Question: Does the paper report error bars suitably and correctly defined or other appropriate information about the statistical significance of the experiments?
    \item[] Answer: \crhl{\answerYes{}}
    \item[] Justification: We follow the standard evaluation protocol of each benchmark (e.g., 500 rollouts for LIBERO, 1000 chains for CALVIN).

\item {\bf Experiments compute resources}
    \item[] Question: For each experiment, does the paper provide sufficient information on the computer resources (type of compute workers, memory, time of execution) needed to reproduce the experiments?
    \item[] Answer: \answerYes{}
    \item[] Justification: The Appendix (Training Hyperparameters) specifies GPU type (H100), number of GPUs (4 for simulation, 64 for real-robot), batch sizes, and training steps. Table~\ref{tab:efficiency} reports peak VRAM usage.

\item {\bf Code of ethics}
    \item[] Question: Does the research conducted in the paper conform, in every respect, with the NeurIPS Code of Ethics \url{https://neurips.cc/public/EthicsGuidelines}?
    \item[] Answer: \answerYes{}
    \item[] Justification: The research involves robotic manipulation in controlled lab and simulation settings. No human subjects, private data, or dual-use concerns are involved.

\item {\bf Broader impacts}
    \item[] Question: Does the paper discuss both potential positive societal impacts and negative societal impacts of the work performed?
    \item[] Answer: \answerYes{}
    \item[] Justification: The Broader Impact subsection discusses positive impacts (reproducibility, accessibility, and reduced research cost) and potential risks (safety validation, human oversight, and labor impacts) of stronger robotic manipulation policies.

\item {\bf Safeguards}
    \item[] Question: Does the paper describe safeguards that have been put in place for responsible release of data or models that have a high risk for misuse (e.g., pre-trained language models, image generators, or scraped datasets)?
    \item[] Answer: \answerNA{}
    \item[] Justification: SimVLA is a robotic manipulation policy evaluated in controlled settings. The model does not pose risks comparable to generative language or image models.

\item {\bf Licenses for existing assets}
    \item[] Question: Are the creators or original owners of assets (e.g., code, data, models), used in the paper, properly credited and are the license and terms of use explicitly mentioned and properly respected?
    \item[] Answer: \answerYes{}
    \item[] Justification: All benchmarks (LIBERO, CALVIN, WidowX, Google Robot, BEHAVIOR-1K), baseline models, and the SmolVLM, Florence-2, and Qwen3-VL backbones are cited with their original papers.

\item {\bf New assets}
    \item[] Question: Are new assets introduced in the paper well documented and is the documentation provided alongside the assets?
    \item[] Answer: \answerYes{}
    \item[] Justification: We release the SimVLA codebase with documentation at \url{https://frontierrobo.github.io/SimVLA/}.

\item {\bf Crowdsourcing and research with human subjects}
    \item[] Question: For crowdsourcing experiments and research with human subjects, does the paper include the full text of instructions given to participants and screenshots, if applicable, as well as details about compensation (if any)?
    \item[] Answer: \answerNA{}
    \item[] Justification: This work does not involve crowdsourcing or research with human subjects.

\item {\bf Institutional review board (IRB) approvals or equivalent for research with human subjects}
    \item[] Question: Does the paper describe potential risks incurred by study participants, whether such risks were disclosed to the subjects, and whether Institutional Review Board (IRB) approvals (or an equivalent approval/review based on the requirements of your country or institution) were obtained?
    \item[] Answer: \answerNA{}
    \item[] Justification: This work does not involve human subjects research.

\item {\bf Declaration of LLM usage}
    \item[] Question: Does the paper describe the usage of LLMs if it is an important, original, or non-standard component of the core methods in this research? Note that if the LLM is used only for writing, editing, or formatting purposes and does \emph{not} impact the core methodology, scientific rigor, or originality of the research, declaration is not required.
    \item[] Answer: \answerNA{}
    \item[] Justification: LLMs are not used as a component of the core methodology. 

\end{enumerate}

\end{document}